\documentclass[10pt,twocolumn,letterpaper]{article}

\usepackage{iccv}
\usepackage{times}
\usepackage{epsfig}
\usepackage{graphicx}
\usepackage{amsmath}
\usepackage{amssymb}

\usepackage{subfigure}
\usepackage{cite}
\usepackage{float}
\usepackage{bm}

\usepackage[breaklinks=true,bookmarks=false]{hyperref}
\hypersetup{
	colorlinks=true,
	linkcolor=blue,
	filecolor=blue,      
	urlcolor=blue,
	citecolor=cyan,
}

\iccvfinalcopy 

\def\iccvPaperID{6944} 

\ificcvfinal\fi
\begin{document}

\title{Learning Aberrance Repressed Correlation Filters for Real-Time UAV Tracking}

\author{Ziyuan Huang${^1}$, Changhong Fu$^{2,*}$, Yiming Li${^2}$, Fuling Lin${^2}$ and Peng Lu${^3}$\\
${^1}$School of Automotive Studies,
${^2}$School of Mechanical Engineering,
Tongji University,
China\\
${^3}$Adaptive Robotic Controls Lab, Hong Kong Polytechnic University, Hong Kong, China\\
{\tt\small tjhuangziyuan@gmail.com, changhongfu@tongji.edu.cn, peng.lu@polyu.edu.hk}
}

\maketitle
\ificcvfinal\thispagestyle{empty}\fi

\begin{abstract}
	Traditional framework of discriminative correlation filters (DCF) is often subject to undesired boundary effects. Several approaches to enlarge search regions have been already proposed in the past years to make up for this shortcoming. However, with excessive background information, more background noises are also introduced and the discriminative filter is prone to learn from the ambiance rather than the object. This situation, along with appearance changes of objects caused by full/partial occlusion, illumination variation, and other reasons has made it more likely to have aberrances in the detection process, which could substantially degrade the credibility of its result. Therefore, in this work, a novel approach to repress the aberrances happening during the detection process is proposed, i.e., aberrance repressed correlation filter (ARCF). By enforcing restriction to the rate of alteration in response maps generated in the detection phase, the ARCF tracker can evidently suppress aberrances and is thus more robust and accurate to track objects. Considerable experiments are conducted on different UAV datasets to perform object tracking from an aerial view, i.e., UAV123, UAVDT, and DTB70, with 243 challenging image sequences containing over 90K frames to verify the performance of the ARCF tracker and it has proven itself to have outperformed other 20 state-of-the-art trackers based on DCF and deep-based frameworks with sufficient speed for real-time applications.
\end{abstract}

\section{Introduction}

Visual object tracking has been widely applied in numerous fields, especially in unmanned aerial vehicle (UAV) applications, where it has been used for target following~\cite{Cheng2017IROS}, mid-air aircraft tracking~\cite{Fu2014ICRA} and aerial refueling~\cite{Yin2016TIM}. Due to fast motion of both UAV and tracked object, occlusion, deformation, illumination variation, and other challenges, robust and accurate tracking has remained a demanding task. 
\begin{figure}[t]
	\begin{center}
		\includegraphics[width=1\linewidth]{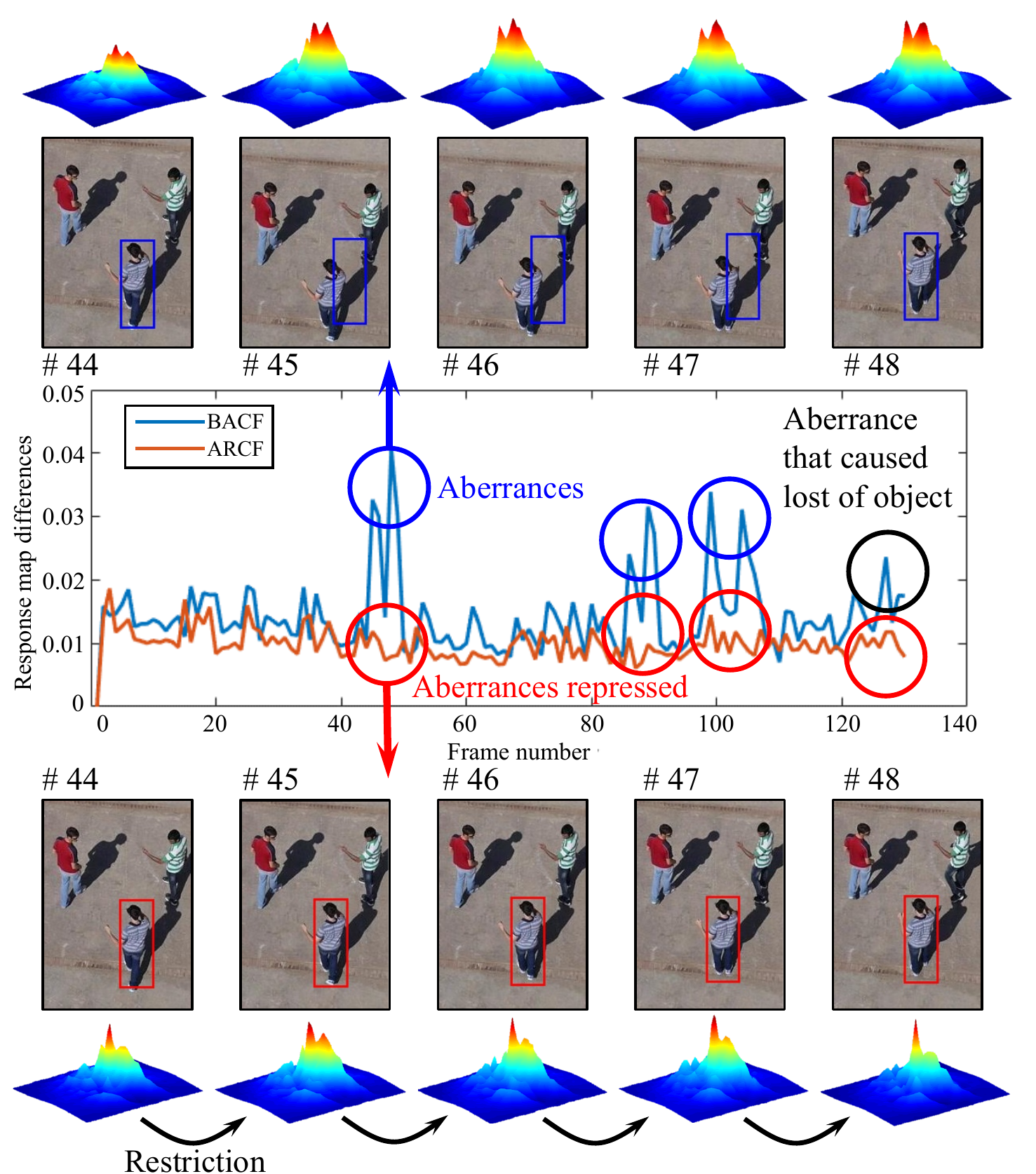}
	\end{center}
	\caption{Comparison between background-aware correlation filter (BACF) and the proposed ARCF tracker. The central figure is to demonstrate the differences between previous response map and current response map on \textit{group1\_1} from UAV123@10fps. Sudden changes of response maps indicate aberrances. When aberrances take place, BACF is tend to lose track of the object while the proposed ARCF can repress aberrances so that this kind of drifting can be avoided.}
\end{figure}

In recent years, discriminative correlation filter (DCF) has contributed tremendously to the field of visual tracking because of its high computational efficiency. It utilizes a property of circulant matrices to carry out the otherwise complicated calculation in the frequency domain rather than in the spatial domain to raise computing speed. Unfortunately, utilization of this property creates artificial samples, leading to undesired boundary effects, which severely degrades tracking performances.

In detection process, traditional DCF framework generates a response map and the object is believed to be located where its value is the largest. Information hidden in response map is crucial as its quality to some extent reflects the similarity between object appearance model learned in previous frames and the actual object detected in current frame. Aberrances are omnipresent in occlusion, in/out of the plan rotation and many other challenging scenarios. However, traditional DCF framework fails to utilize this information and when aberrances take place, no action can be further taken and the tracked object is simply lost.

In UAV object tracking, these two problems are especially crucial. There are relatively more cases of fast motion or low resolution and lack of search region can thus easily result in drift or lost of object. Objects also go through more out-of-the-plane rotations and thus aberrances are more likely to take place in aerial tracking scenarios. In addition, with restricted calculate capability, a tracker that can cope with these two problems and perform efficiently is especially needed. 

\subsection{Main contributions}
This work proposes a novel tracking approach that resolves both aforementioned problems, i.e., ARCF tracker. A cropping matrix and a regularization term are introduced respectively for search region enlargement and for aberrance repression. An efficient convex optimization method is applied in order to ensure sufficient computing efficiency.

Contributions of this work can be listed as follows:
\begin{itemize}
\item A novel tracking method capable of effectively and efficiently suppressing aberrances while solving boundary effects is proposed. Background patches are fed into both learning and detection process to act as negative training samples and to enlarge search areas. A regularization term to restrict the change rate of response maps is added so that abrupt alteration of response maps can be avoided.
\item The proposed ARCF tracker is exhaustively tested on 243 challenging image sequences captured by UAV. Both hand-crafted based trackers, i.e., histogram of oriented gradient (HOG) and color names (CN), and deep trackers are compared in the extensive experiments with the proposed ARCF tracker. Thorough evaluations have demonstrated that ARCF tracker performs favorably against other 20 state-of-the-art trackers.
\end{itemize}

To the best of our knowledge, this is the first time aberrance repression formulation has been applied in DCF framework. It can raise the robustness of DCF based trackers and improve their performances in UAV tracking tasks.

\section{Related work}
\subsection{Discriminative correlation filter}
Discriminative correlation filter based framework has been broadly applied to visual tracking since it was first introduced by Bolme \etal~\cite{Bolme2010CVPR} who proposed a method called minimum output sum of squared error (MOSSE) filter. Kernel trick was introduced to DCF framework by Henriques \etal~\cite{Henriques2015PAMI} to achieve better performance. Introduction of scale estimation has further improved the framework~\cite{Li2014ECCV}. Context and background information are also exploited to have more negative samples so that learned correlation filters can have more discriminative power\cite{Mueller2017CVPR,Kiani2017ICCV,Danelljan2015ICCV}. Besides hand-crafted features used in~\cite{Li2014ECCV,Kiani2017ICCV,Henriques2015PAMI,Danelljan2015ICCV}, the application of deep features is also investigated for more precise and comprehensive object appearance representation~\cite{Ma2015ICCV,Danelljan2015ICCVWorkshop, Fu2019IROS}. Some trackers combine hand-crafted features with deep ones to better describe the tracked objects from multiple aspects~\cite{Li2018CVPR, Danelljan2017CVPR}. DCF based trackers have achieved state-of-the-art performance in multiple datasets specified for UAV object tracking~\cite{Du2018ECCV,Li2017AAAI,Mueller2016ECCV}.

\subsection{Prior solution to boundary effects}
As was stated before, traditional DCF based framework usually suffers from boundary effects due to the limited search region originating from its periodic shifting of the area near original object. Some measures are already taken to mitigate this effect~\cite{Kiani2017ICCV,Danelljan2015ICCV, Fu2019IROS}. Spatially regularized DCF (SRDCF) was proposed to introduce punishment for background in training correlation filters so that they can be learned in larger search regions~\cite{Danelljan2015ICCV}. Unfortunately, this method has high computational costs. Background-aware correlation filter (BACF) extracts patches densely from background using cropping matrix~\cite{Kiani2017ICCV}, which expands search region with lower computational cost. Background effect-aware visual tracker (BEVT) merges these two methods, thus achieving a better performance~\cite{Fu2019IROS}. 
\begin{figure*}[!t]
	\begin{center}
		\includegraphics[width=1\linewidth]{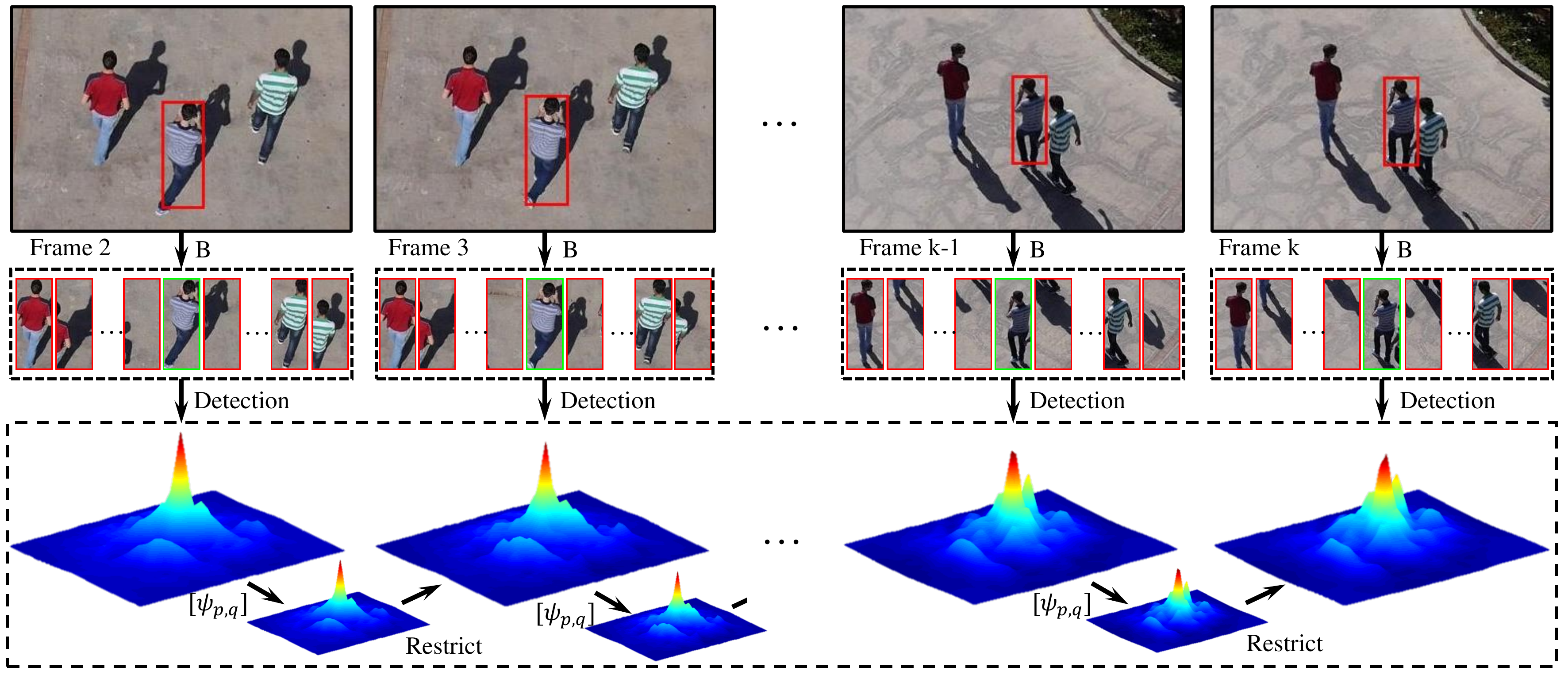}
	\end{center}
	\caption{Main work-flow of the proposed ARCF tracker. It learns both positive sample (green samples) of the object and negative samples (red samples) extracted from the background and the response map restriction is integrated in the learning process so that aberrances in response maps can be repressed. $[\psi_{p,q}]$ serves to shift the generated response map so that the peak position in the previous frame is the same as that of the current frame and thus the position of the detected object will not affect the restriction.}
	\label{fig:overall_structure}
\end{figure*}
\subsection{Prior solution to aberrances}
There is few attention paid to information revealed in response maps. Wang \etal proposed a method called LMCF where the quality of response maps is verified in the learning phase and used to perform high-confidence update of appearance models~\cite{Wang2017CVPR}, which reduces the learning rate to zero in low-confidence situations. Attentional correlation filter network (ACFN) integrates a few trackers as a network and generates a validation score for response maps from each frame. A neural network is trained based on that score to choose a suitable tracker in the next frame~\cite{Choi2017CVPR}. However, both methods take measures after the possible aberrances, which can only have limited influence in suppressing those aberrances compared to the proposed ARCF tracker which tries to repress aberrances during the training phase.

\section{Background-aware correlation filter}
\label{sec:BACF}
In this section, background-aware correlation filter (BACF)~\cite{Kiani2017ICCV}, on which our method is based, is reviewed. 

Given the vectorized sample $\mathbf{x}$ with $D$ channels of $\mathbf{x}^d\in\mathbb{R}^N(d=1,2,...,D)$ and the vectorized ideal response $\mathbf{y}\in\mathbb{R}^N$, the overall objective of BACF is to minimize the objective $\mathcal { E } ( \mathbf { w } )$, i.e.,
\begin{equation}\label{eq:bacf}
\mathcal { E } ( \mathbf { w } ) = \frac { 1 } { 2 } \Vert \mathbf { y } - \sum _ { d = 1 } ^ { D } \mathbf { Bx }^d \star \mathbf { w } ^ { d } \Vert _2^2 + \sum _ { d = 1 } ^ { D } \Vert \mathbf { w } ^ { d } \Vert _2^2 \ ,
\end{equation}
\noindent where $\mathbf{B} \in \mathbb{R}^{M\times N}$ is a cropping matrix to select central $M$ elements of the each channel $\mathbf{x}^d$ of input vectorized sample, and $\mathbf{w}^{d}\in\mathbb{R}^M$ is the correlation filter to be learned in $d$-th channel. Usually, $M<<N$. The operator $\star$ is a correlation operator.

By introducing the cropping matrix, BACF is able to utilize not only objects but also real background information instead of shifted patches in training process of correlation filters. Due to expanded search region, it is capable of tracking an object with relatively high relative speed to the camera or UAV. Unfortunately, with excessive background information, more background clutter is introduced and similar objects in the contexts are more likely than prior DCF frameworks to be detected and recognized as the original object being tracked. When this problem is observed in the response map, it can be clearly seen that BACF does not handle well when aberrances take place.

\section{Aberrance repressed correlation filter}
As stated in \ref{sec:BACF}, BACF, just as other DCF based trackers, is vulnerable when aberrance happens. In this work, an aberrance repressed correlation filter, i.e., ARCF, is proposed to suppress sudden changes of response maps. The main structure can be seen in Fig.~\ref{fig:overall_structure}.
\subsection{Overall objective of ARCF}
Compared to other measures taken after occurences of aberrances in LMCF and ACFN, the proposed ARCF tracker aims to integrate the suppression of their occurences to the trainig process of correlation filters. In order to repress aberrances, they should be firstly identified. Euclidean norm is introduced to define difference level of two response maps $\mathbf{M}_1$ and $\mathbf{M}_2$ as follows:
\begin{equation}\label{eq:difference}
\Vert \mathbf{M}_1[\psi_{p,q}] - \mathbf{M}_2 \Vert _2^2 \ ,
\end{equation}
\noindent where $p$ and $q$ denote the location difference of two peaks in both response maps in two-dimensional space and $[\psi_{p,q}]$ indicates the shifting operation in order for two peaks to coincide with each other. Usually when an aberrance takes place, the similarity would suddenly drop and thus the value of Eq.~\ref{eq:difference} will be high. By judging the value of Eq.~\ref{eq:difference}, the aberrances can easily be identified.

In order to repress aberrances in the training process, the training objective is optimized to minimize the loss function as follows:
\begin{equation}\label{eq:objective}
\begin{small}
\begin{aligned}
& \mathcal { E } ( \mathbf { w }_k ) = \frac { 1 } { 2 } \Vert \mathbf { y } - \sum_{ d=1 }^{ D } \mathbf { Bx }_k^d \star \mathbf { w }_k^d\Vert _2^2 + \frac{\lambda}{2}\sum_{ d=1 }^{ D } \Vert \mathbf { w }_k^d \Vert _2^2 \\
& + \frac{\gamma}{2} \Vert \sum_{ d=1 }^{ D } (\mathbf{ Bx }_{k-1}^d \star \mathbf{ w }_{k-1}^d)[\psi_{p,q}] - \sum_{ d=1 }^{ D } \mathbf{ Bx }_k^d \star \mathbf{w}_k^d \Vert_2^2
\end{aligned} \ ,
\end{small}
\end{equation}
\noindent where subscript $k$ and $k-1$ denote the $k$th and $(k-1)$th frame respectively. The third term of Eq.~\ref{eq:objective} is a regularization term to restrict the aberrances mentioned before. Parameter $\gamma$ is introduced as the aberrance penalty. In the following transformation and optimization part, the restriction will be transformed into frequency domain and optimized so that the repression can be carried out in the training process of correlation filters. 

Here the cropping matrix $\mathbf{ B }$ is retained from BACF to ensure sufficient search region. Meanwhile, the regularization term is introduced to counteract the aberrances that background information has brought by expanding search area. 

In order for the overall objective to be more easily transformed into frequency domain, it is firstly expressed in matrix form as follows:
\begin{equation}\label{eq:objective_matrix}
\begin{aligned}
\mathcal { E } ( \mathbf { w }_k ) 
& = \frac { 1 } { 2 } \Vert \mathbf { y } - \mathbf{ X }_k \left( \mathbf{ I }_D \otimes \mathbf{ B }^{\top} \right) \mathbf{ w }_k \Vert _2^2 + \frac{\lambda}{2} \Vert \mathbf { w }_k \Vert _2^2 \\
& + \frac{\gamma}{2} \Vert \mathbf{M}_{k-1}[\psi_{p,q}] - \mathbf{ X }_{k} \left( \mathbf{ I }_D \otimes \mathbf{ B }^{\top} \right) \mathbf{ w }_k \Vert_2^2
\end{aligned} \ ,
\end{equation}
\noindent where $\mathbf{X}_{k}$ is the matrix form of input sample $\mathbf{x}_{k}$. $\mathbf{I}_D$ is an identity matrix whose size is $D \times D$. Operator $\otimes$ and superscript $\top$ indicates respectively Kronecker production and conjugate transpose operation.  $\mathbf{M}_{k-1}$ denotes the response map from previous frame and its value is equivalent to $\mathbf{ X }_{k-1} \left( \mathbf{ I }_D \otimes \mathbf{ B }^{\top} \right) \mathbf{ w }_{k-1}$.

\subsection{Transformation into frequency domain}
Although the overall loss function can be expressed in matrix form as Eq.~\ref{eq:objective_matrix}, essentially it is still carrying out convolution operation. Therefore, to minimize the overall objective, Eq.~\ref{eq:objective_matrix} is also transformed into frequency domain as follows to ensure sufficient computing efficiency:
\begin{equation}\label{eq:frequency_format}
\begin{aligned}
&\mathcal { \hat{E} } ( \mathbf { w }_k, \mathbf{\hat{g}}_k ) = \frac{1}{2}  \lVert \mathbf{\hat{y}} 
- \mathbf{\hat{X}}_k \mathbf{\hat{g}}_k \rVert_2^2 + 
\frac{\lambda}{2} \lVert \mathbf{w}_k \rVert _2^2
\\
&\quad\quad\quad\quad + \frac{\gamma}{2}\Vert \mathbf{\hat{M}}^s_{k-1} - \mathbf{ \hat{X} }_{k}\mathbf{\hat{g}}_k \Vert_2^2
\\
& \quad \quad \text{s.t.}\quad \hat{\mathbf{g}}_k = \sqrt{N}(\mathbf{I}_D \otimes \mathbf{F} \mathbf{B}^{\top} )\mathbf{w}_k
\end{aligned} \ ,
\end{equation}
\noindent where the superscript $ \hat{} $ denotes a signal that has been performed discrete Fourier transformation (DFT), i.e., $\hat{\bm{\alpha}} = \sqrt{N}\mathbf{F}\bm{\alpha}$. A new parameter $\mathbf{\hat{g}}_k \in \mathbb{C} ^{DN \times 1}$ is introduced in preparation for further optimization. $\mathbf{\hat{M}}_{k-1}^s$ denotes the discrete Fourier transformation of shifted signal $\mathbf{M}_{k-1}[\psi_{p,q}]$. Since in the current frame, the response map in the former frame is already generated, $\mathbf{\hat{M}}_{k-1}^s$ can be treated as a constant signal, which can simplify the further calculation.  

\subsection{Optimization through ADMM}
Similar to BACF tracker, alternative direction method of multipliers (ADMM) is applied to speed up calculation. Due to the convexity of equation \ref{eq:frequency_format}, it can be minimized using ADMM to achieve a global optimal solution. Therefore, Eq.~\ref{eq:frequency_format} is first required to be written in augmented Lagrangian form as follows:
\begin{equation}\label{eq:augmented_lagrangian}
\begin{aligned}
&\mathcal { \hat{E} } ( \mathbf { w }_k, \mathbf{\hat{g}}_k, \hat{\bm{\zeta}} ) = \frac{1}{2}  \lVert \mathbf{\hat{y}} 
- \mathbf{\hat{X}}_k \mathbf{\hat{g}}_k \rVert_2^2 + 
\frac{\lambda}{2} \lVert \mathbf{w}_k \rVert _2^2
\\
&\quad\quad + \frac{\gamma}{2}\Vert \mathbf{\hat{M}}^s_{k-1} - \mathbf{ \hat{X} }_{k}\mathbf{\hat{g}}_k \Vert_2^2
\\
&\quad\quad + \hat{\bm{\zeta}}^{\top} \left( \mathbf{\hat{g}}_k - \sqrt{N}(\mathbf{I}_D \otimes \mathbf{F} \mathbf{B}^{\top} )\mathbf{w}_k \right)
\\
&\quad\quad + \frac{\mu}{2} \Vert \mathbf{\hat{g}}_k - \sqrt{N}(\mathbf{I}_D \otimes \mathbf{F} \mathbf{B}^{\top} )\mathbf{w}_k \Vert _2^2
\end{aligned} \ ,
\end{equation}
\noindent where $\mu$ is introduced as a penalty factor and the Lagrangian vector in the Fourier domain $\hat{\bm{\zeta}} = [\hat{\bm{\zeta}}^{1\top}, \cdots, \hat{\bm{\zeta}}^{D\top}]^\top$ is introduced as auxiliary variable that has a size of ${DN\times 1}$. 

Employing ADMM in the $k$th frame means that the augmented Lagrangian form can be solved by solving two subproblems, respectively the following $\mathbf{w}_{k+1}^*$ and $\mathbf{\hat{g}}_{k+1}^*$ to calculate correlation filters for the $(k+1)$th frame:
\begin{equation}\label{eq:subproblems}
\begin{small}
\begin{cases}
\begin{aligned}
\mathbf{w}_{k+1}^*=
&\arg\min _ { \mathbf { w }_k } \Big\lbrace
\frac{\lambda}{2}\lVert \mathbf{w}_k \rVert _2^2 
\\
&+ \hat{\bm{\zeta}}^{\top}\left(\hat{\mathbf{g}_k} - \sqrt{N} \left(\mathbf{I}_D \otimes \mathbf{FB}^{\top}\right)\mathbf{w}_k\right)
\\
&+ \frac{\mu}{2} \lVert \hat{\mathbf{g}}_k - \sqrt{N} \left(\mathbf{I}_D \otimes \mathbf{FB}^{\top}\right) \mathbf{w}_k \rVert _2^2
\Big\rbrace
\end{aligned}
\\
\begin{aligned}
\hat{\mathbf{g}}_{k+1}^* = &\arg \min_{ \mathbf { g }_k }  \Big\lbrace
\frac{1}{2}  \lVert \hat{\mathbf{y}} - 
\hat {\mathbf{X}}_k \hat{\mathbf{g}}_k\rVert_2^2 
\\
&+ \frac{\gamma}{2}\Vert \mathbf{\hat{M}}^s_{k-1} - \hat {\mathbf{X}}_k \hat{\mathbf{g}}_k \Vert_2^2
\\
&+ \hat{\bm{\zeta}}^{\top} \left(\hat{\mathbf{g}}_k - \sqrt{N} \left(\mathbf{I}_D \otimes \mathbf{FB}^{\top}\right)\mathbf{w}_k \right)
\\
&+ \frac{\mu}{2} \lVert \hat{\mathbf{g}}_k - \sqrt{N} \left(\mathbf{I}_D \otimes \mathbf{FB}^{\top}\right) \mathbf{w}_k \rVert _2^2
\Big\rbrace
\end{aligned}
\end{cases}
\end{small}
 \ .
\end{equation}
Both of these two subproblems have closed-form solutions.

\subsubsection{Solution to subproblem $\mathbf{w}_k^*$}
The solution to subproblem $\mathbf{w}_k^*$ can be easily obtained as follows:
\begin{equation}
\begin{small}
\begin{aligned}
\mathbf{w}_{k+1}^* & = \left( \lambda + \mu N \right)^{-1}
\\
& \left(\sqrt{N} \left(\mathbf { I } _ { D } \otimes \mathbf { BF } ^ { \top } \right)\hat{\bm{\zeta}} + \mu\sqrt{N}\left(\mathbf { I } _ { D } \otimes \mathbf { BF } ^ { \top } \right)\mathbf{\hat{g}}_k \right)
\\
& =\left(\frac{\lambda}{N} + \mu\right)^{-1} \left(\bm{\zeta} + \mu \mathbf{g}_k\right)
\end{aligned}
\end{small}
 ,
\end{equation}
\noindent where $\mathbf{g}_k$ and $\bm{\zeta}$ can be obtained respectively through following inverse fast Fourier transformation operations:
\begin{equation}
\begin{cases}
\begin{aligned}
\mathbf{g}_k & = \frac { 1 } { \sqrt { N } } \left(\mathbf { I } _ { D } \otimes \mathbf { BF } ^ { \top } \right) \hat { \mathbf { g } }_k \\
\mathbf { \bm{\zeta} } & = \frac { 1 } { \sqrt { N } } \left(  \mathbf { I } _ { D } \otimes \mathbf { BF } ^ { \top } \right) \hat { \mathbf { \bm{\zeta} } } \\ 
\end{aligned}
\end{cases}\ .
\end{equation}

\subsubsection{Solution to subproblem $\mathbf{\hat{g}}_k^*$}
Unfortunately, unlike subproblem $\mathbf{w}_k^*$, solving subproblem $\mathbf{\hat{g}}_k^*$ containing $\hat {\mathbf{X}}_k \hat{\mathbf{g}}_k$ can be highly time consuming and the calculation needs to be carried out in every ADMM iteration. Therefore, the sparsity of $\hat {\mathbf{X}}_k$ is exploited. Each element of $\mathbf{\hat{y}}$, i.e., $\mathbf{\hat{y}}(n), n=1,2,...,N,$, is solely dependent on each $\hat{\mathbf{x}}_k(n)=\left[\hat {\mathbf{x}}_k^1(n),\hat {\mathbf{x}}_k^2(n),...,\hat {\mathbf{x}}^D(n)\right]^{\top}$ and $\hat { \mathbf { g } }_k ( n ) = \left[ \operatorname { conj } \left( \hat { \mathbf { g } }_k ^ { 1 } ( n ) \right) , \ldots , \operatorname { conj } \left( \hat { \mathbf { g } }_k ^ { D } ( n ) \right) \right] ^ { \top }$. Operator $\operatorname{conj(.)}$ denotes the complex conjugate operation. 

The subproblem $\mathbf{\hat{g}}_k^*$ can be thus further divided into $N$ smaller problems as follows solved over $n=[1,2,...N]$:
\begin{equation}
\begin{aligned}
\hat{\mathbf{g}}_{k+1}(n)^* = &\arg \min_{ \mathbf { g }_k(n) }  \Big\lbrace
\frac{1}{2}  \lVert \hat{\mathbf{y}}(n) - 
\hat {\mathbf{x}}^{\top}_k(n) \hat{\mathbf{g}}_k(n)\rVert_2^2 
\\
&+ \frac{\gamma}{2}\Vert \mathbf{\hat{M}}^s_{k-1} - \hat {\mathbf{x}}^{\top}_k(n) \hat{\mathbf{g}}_k(n) \Vert_2^2
\\
&+ \hat{\bm{\zeta}}^{\top}\left(\hat{\mathbf{g}}_k(n) - \mathbf{\hat{w}}_k(n) \right)
\\
&+ \frac{\mu}{2} \lVert \hat{\mathbf{g}}_k(n) - \mathbf{\hat{w}}_k(n) \rVert _2^2
\Big\rbrace
\end{aligned}\ ,
\end{equation}
\noindent where $\hat { \mathbf { w } }_k ( n ) = \left[ \hat { \mathbf { w } }_k ^ { 1 } ( n ) , \ldots , \hat { \mathbf { w } }_k ^ { D } ( n ) \right]$ and $\hat { \mathbf { w } }_k ^ { d }$ is the DFT of $\mathbf { w }_k  ^ { d }$, i.e., $ \hat { \mathbf { w } }_k ^ { d } = \sqrt { D } \mathbf { F } \mathbf { B } ^ { \top } \mathbf { w }_k ^ { d }$. Each smaller problem can be efficiently calculated and solution is presented below:
\begin{equation}\label{eq:solution_g}
\begin{aligned}
& \hat{\mathbf{g}}_{k+1}(n)^* = \frac{1}{1+\gamma}
\left( \hat{\mathbf{x}}_k(n) \hat{\mathbf{x}}_k^{\top}(n) + \frac{\mu}{1+\gamma} \mathbf{I}_D\right)^{-1}
\\
&\left( \hat{\mathbf{x}}_k(n) \hat{\mathbf{y}}(n) + \gamma \hat{\mathbf{x}}_k(n) \mathbf{\hat{M}}^s_{k-1} - \hat{\bm{\zeta}}(n) + \mu \hat{\mathbf{w}}_k(n) \right)
\end{aligned} \ .
\end{equation}

Still, with inverse operation, the calculation can be further optimized and accelerated by applying the Sherman-Morrison formula, i.e., $(\mathbf{A} + \mathbf{u} \mathbf{v}^{\top}) ^ {-1} = \mathbf{A} ^ {-1} - \mathbf{A}^{-1} \mathbf{u} (\mathbf{I}_m + \mathbf{v}^{\top} \mathbf{A}^{-1} \mathbf{u}) ^ {-1} \mathbf{v}^{\top} \mathbf{A}^{-1} $, where $\mathbf{u}$ is an $a\times m$ matrix, $\mathbf{v}$ is and $m \times a$ matrix and $\mathbf{A}$ is an $a\times a$ matrix. In this case, $\mathbf{A}=\frac{\mu}{1+\gamma}\mathbf{I}_D$, and $\mathbf{u} = \mathbf{v} = \mathbf{\hat{x}}_k(n)$. Eq.~\ref{eq:solution_g} is equivalent to the following equation:
\begin{equation}\label{eq:solution_g_sm}
\begin{small}
\begin{aligned}
&\hat{\mathbf{g}}_{k+1}(n)^* 
\\
= &\gamma^{*}
\left( \hat{\mathbf{x}}_k(n) \hat{\mathbf{y}}(n) + \gamma \hat{\mathbf{x}}_k(n) \mathbf{\hat{M}}^s_{k-1} - \hat{\bm{\zeta}}(n) + \mu \hat{\mathbf{w}}_k(n) \right)
\\
- &\gamma^{*}\frac{\hat{\mathbf{x}}_k(n)}{b}
\Big( \hat{\mathbf{S}}_{\mathbf{x}k}(n) \hat{\mathbf{y}}(n) + \gamma \hat{\mathbf{S}}_{\mathbf{x}k}(n) \mathbf{\hat{M}}^s_{k-1} \hat{\mathbf{S}}_{\bm{\zeta}}(n) + \mu \hat{\mathbf{S}}_{\mathbf{w}k}(n) \Big)
\end{aligned}
\end{small}\ ,
\end{equation}

\noindent where $\gamma^{*}=\frac{\mu}{(1+\gamma)^2}$, $\hat{\mathbf{S}}_{\mathbf{x}k}(n)=\hat { \mathbf { x } }_k ( n ) ^{\top} \hat { \mathbf { x } }_k ( n )$, $\hat{\mathbf{S}}_{\bm{\zeta}}(n)=\hat { \mathbf { x } }_k ( n ) ^{\top} \mathbf{\hat{\bm{\zeta}}}$, $\hat{\mathbf{S}}_{\mathbf{w}k}(n)=\hat { \mathbf { x } }_k ( n ) ^{\top} \mathbf{\hat{w}}_k$ and $b=\hat { \mathbf { x } }_k ( n ) ^{\top} \hat { \mathbf { x } }_k ( n ) + \frac{\mu}{1+\gamma}$. Thus far, the subproblems $\mathbf{w}_{k+1}^*$ and $\mathbf{\hat{g}}_{k+1}^*$ are both solved. 

\subsubsection{Update of Lagrangian parameter}
The Lagrangian parameter is updated according to the following equation:
\begin{equation}
\mathbf{\hat{\bm{\zeta}}} ^{(j+1)}_ {k+1} = \mathbf{\hat{\bm{\zeta}}} ^j_ {k+1} + \mu \left(\mathbf{\hat{g}} ^ {*(j+1)} _ {k+1} - \mathbf{\hat{w}} ^ {*(j+1)} _ {k+1} \right) \ , 
\end{equation}
\noindent where the subscript $j$ and $j+1$ denotes the $j$th and the $(j+1)$th iteration respectively. $\mathbf{\hat{g}} ^ {*(j+1)} _ {k+1}$ indicates the solution to the $\mathbf{\hat{g}} ^ {*} _ {k+1}$ subproblem and $\mathbf{\hat{w}} ^ {*(j+1)} _ {k+1}$ indicates the solution to the $\mathbf{w}_{k+1}^*$ subproblem, both in the $(j+1)$th iteration. Here $\mathbf{\hat{w}} ^ {*(j+1)} _ {k+1} = \left(\mathbf{I}_D \otimes \mathbf{FB}^{\top}\right) \mathbf{w} ^ {*(j+1)} _ {k+1}$.
\begin{figure*}[!t]
	\begin{center}

		\subfigure[] { \label{fig:hc_a} 
			\begin{minipage}{0.32\textwidth}
				\centering
				\includegraphics[width=1\columnwidth]{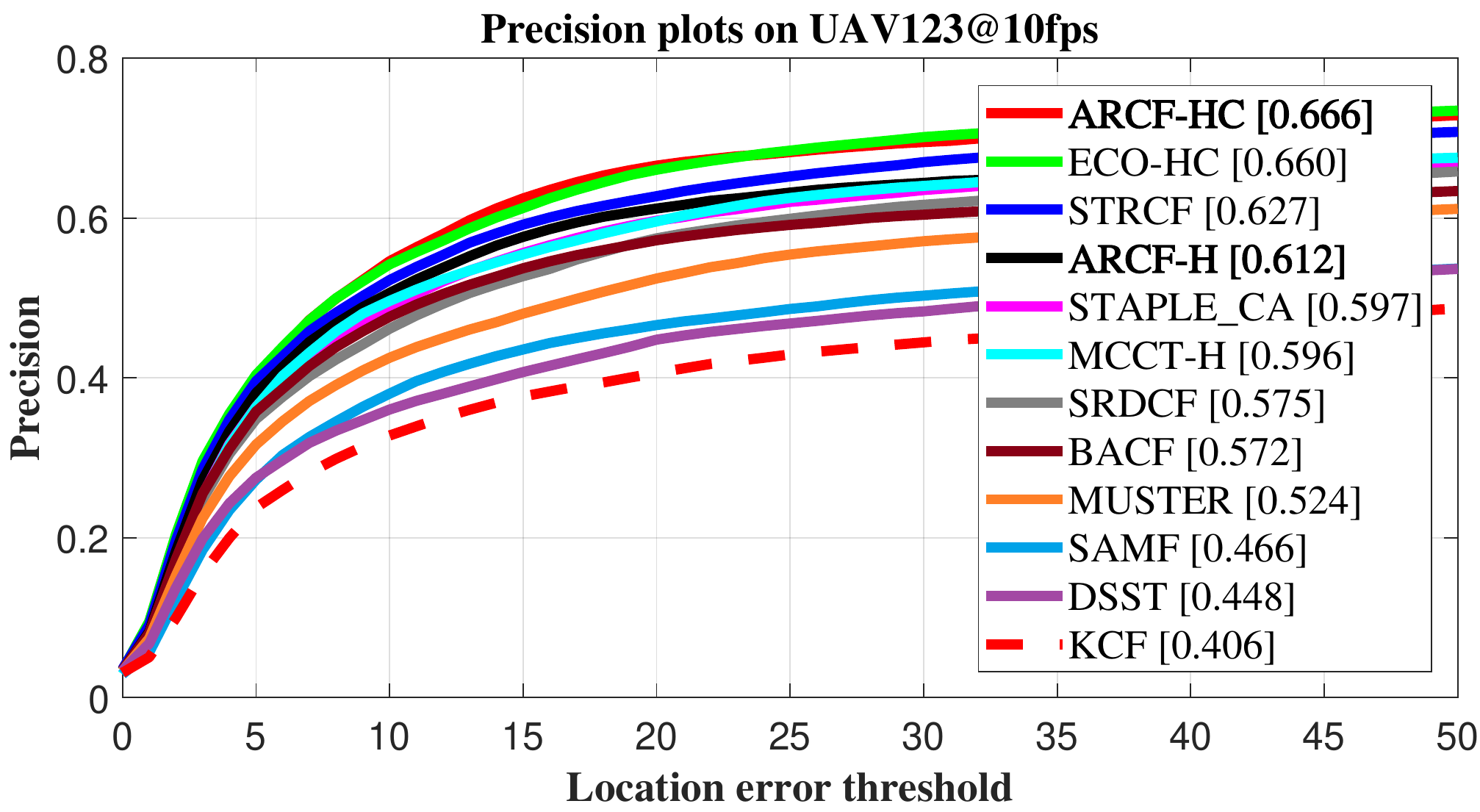}
				\\
				\includegraphics[width=1\columnwidth]{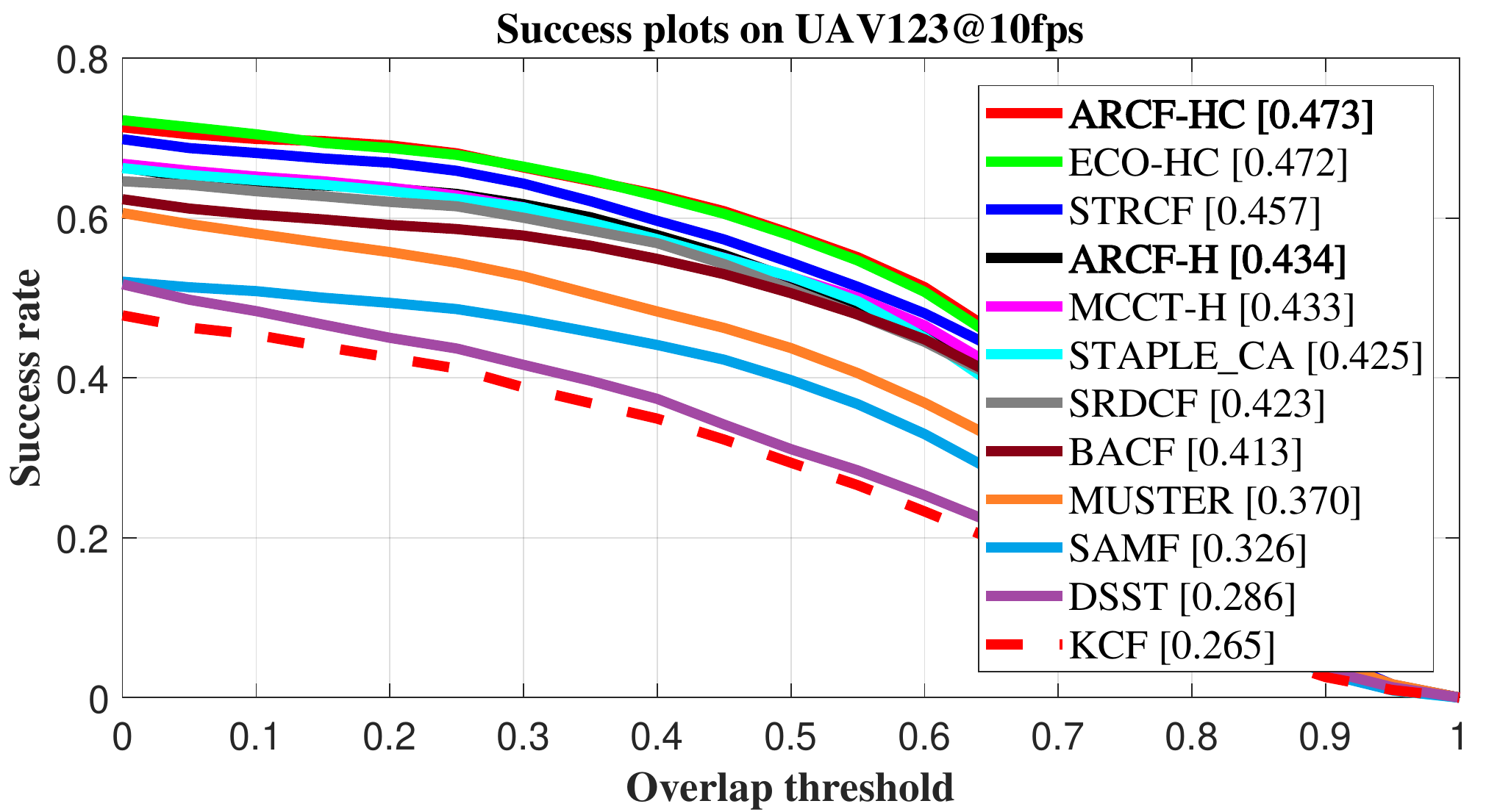}
			\end{minipage}
		}
		\subfigure[] { \label{fig:hc_b} 
			\begin{minipage}{0.32\textwidth}
				\centering
				\includegraphics[width=1\columnwidth]{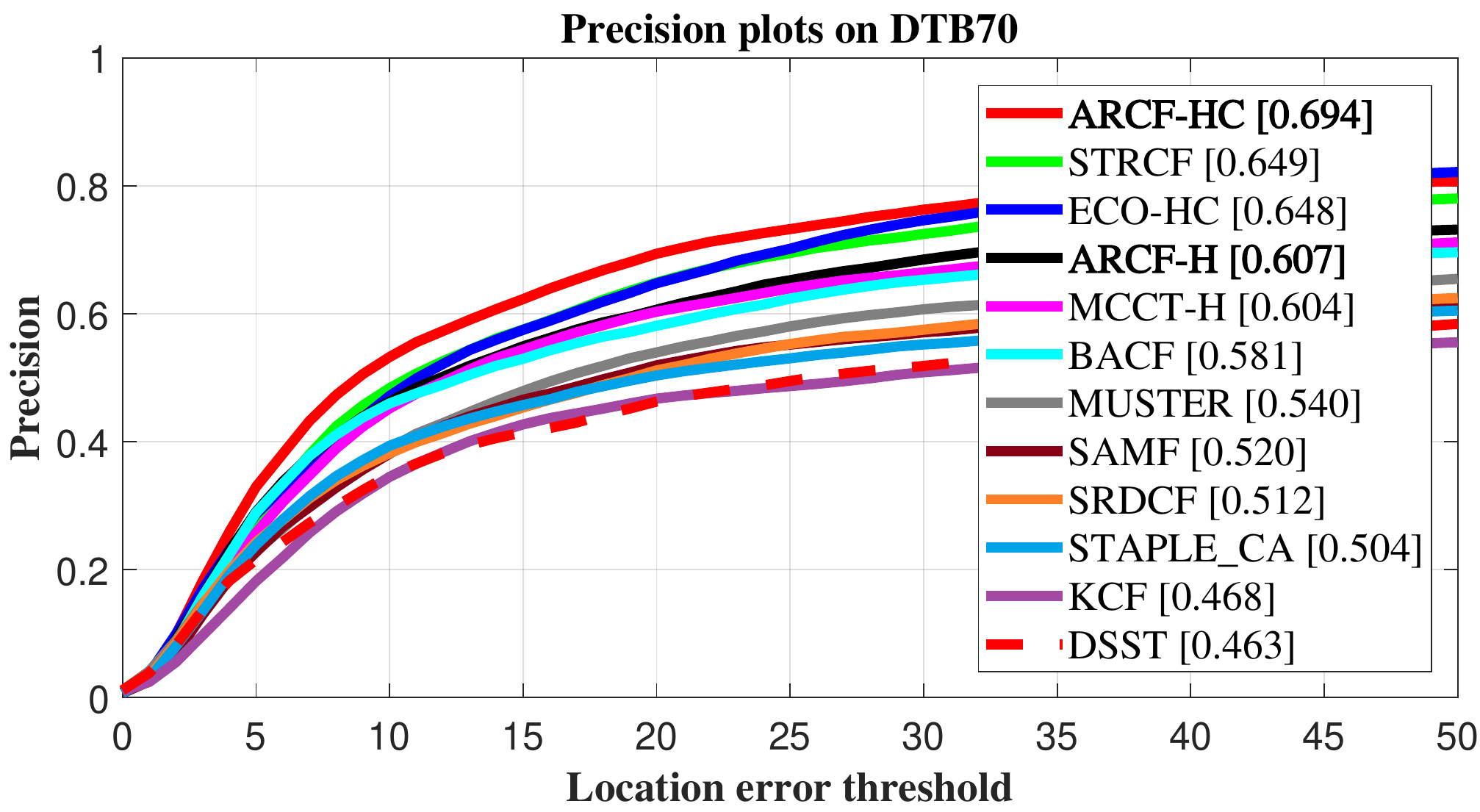}
				\\
				\includegraphics[width=1\columnwidth]{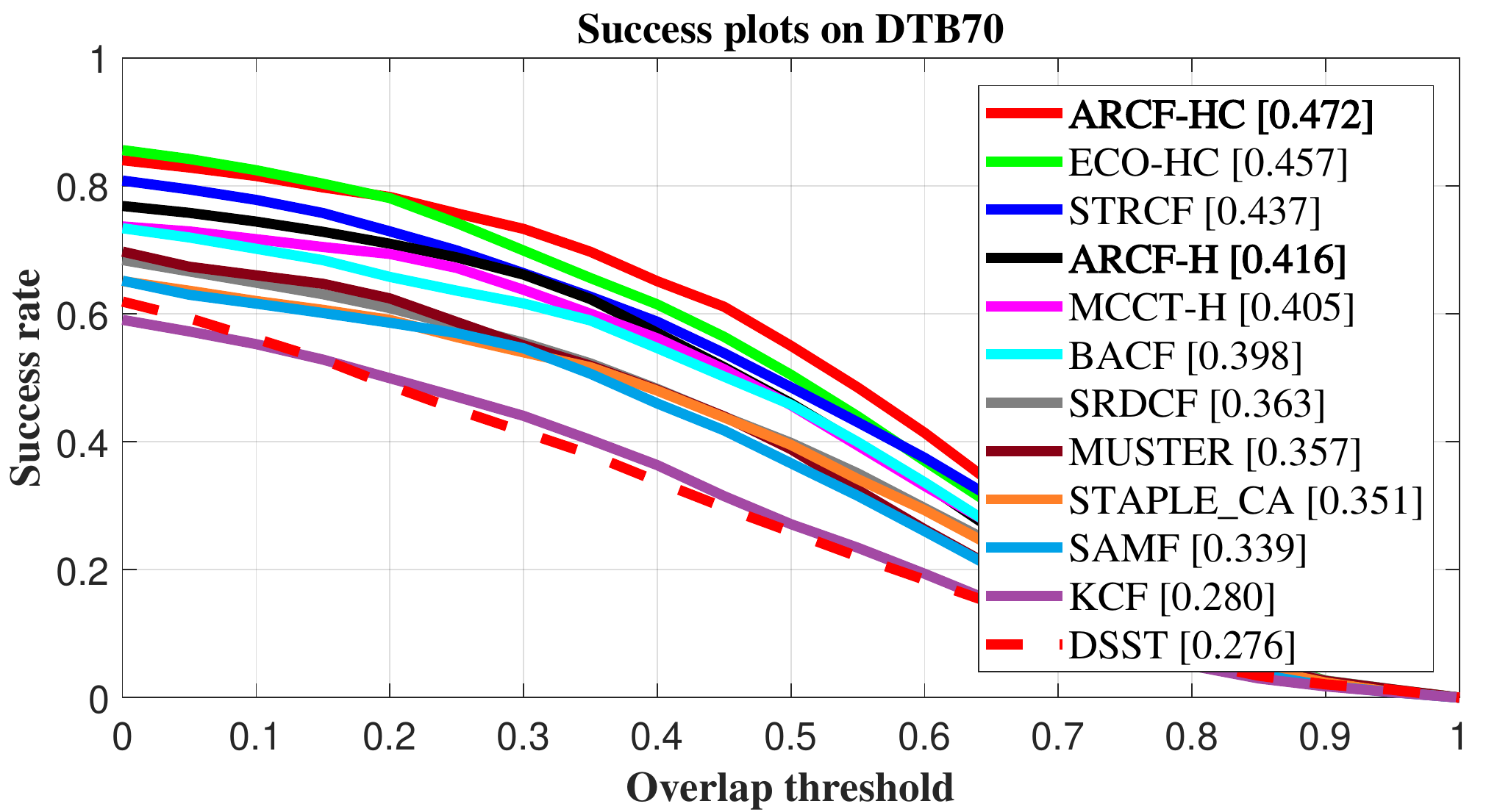}
			\end{minipage}
		}
		\subfigure[] { \label{fig:hc_c} 
			\begin{minipage}{0.32\textwidth}
				\centering
				\includegraphics[width=1\columnwidth]{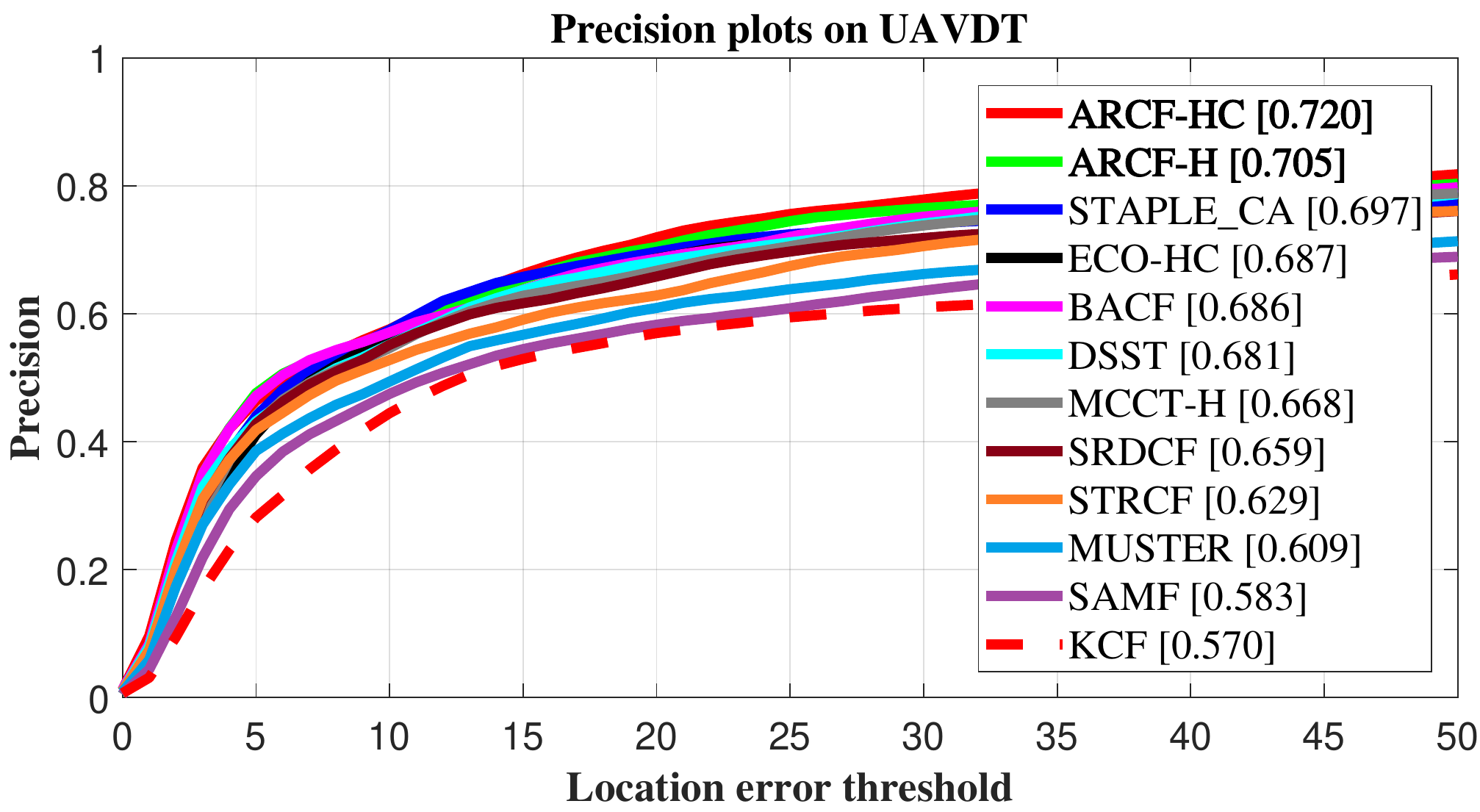}
				\\
				\includegraphics[width=1\columnwidth]{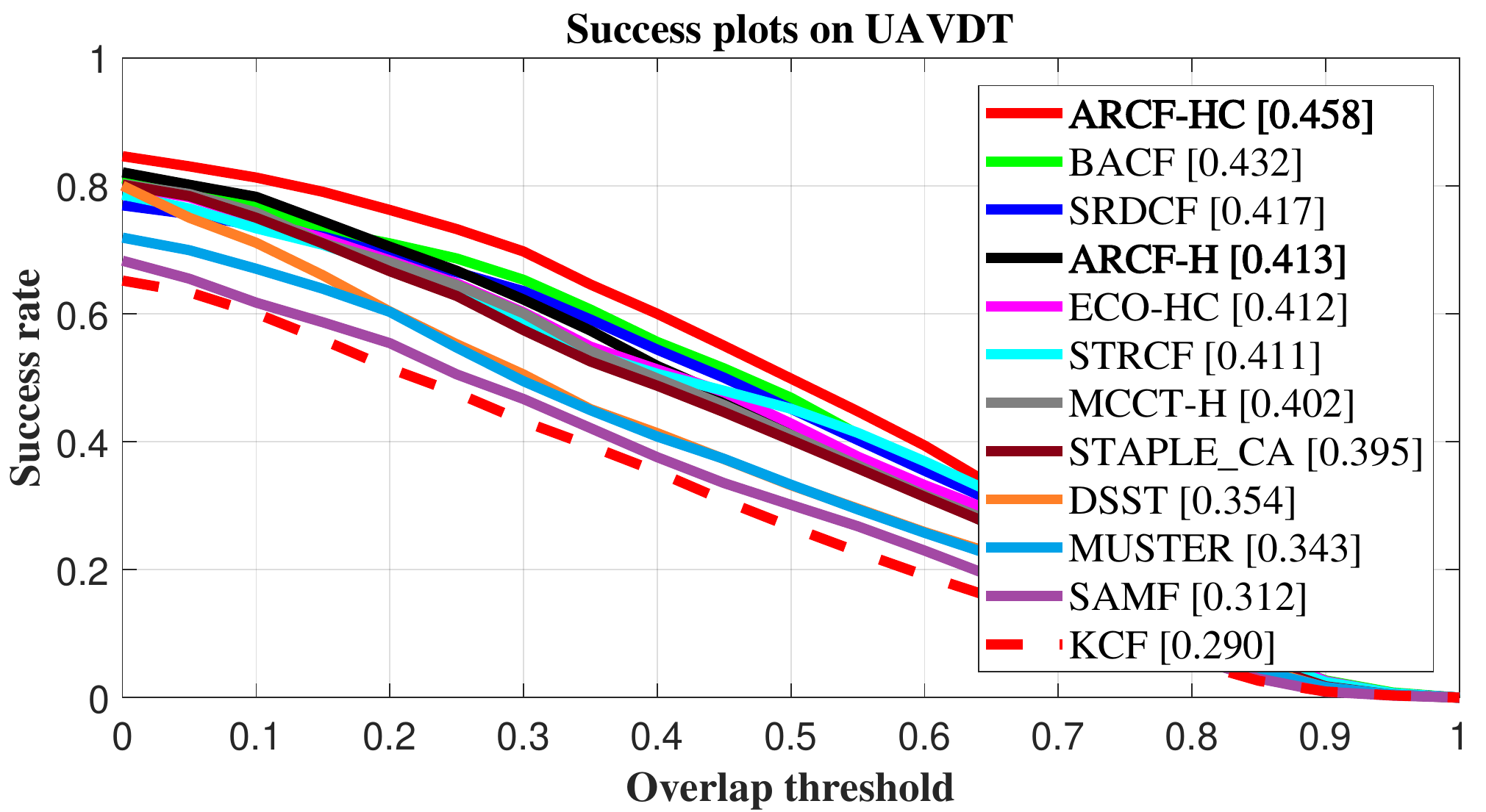}
			\end{minipage}
		}
	\end{center}
	
	\caption{Precision and success plots of ARCF-HC, ARCF-H as well as other hand-crafted feature-based trackers on (a) UAV123, (b) DTB70 and (c) UAVDT. Precision and AUC are marked in the precision plots and success plots respectively. }
	\label{fig:hc}
\end{figure*}
\begin{center}
	\begin{table*}[!t]
		\footnotesize
		\setlength{\tabcolsep}{2mm}
		\centering
		\caption{Average frame per second (FPS) and millisecond per frame (MSPF) of top hand-crafted based trackers on 243 image sequences. \textcolor[rgb]{ 1,  0,  0}{Red}, \textcolor[rgb]{ 0,  1,  0}{green} and \textcolor[rgb]{ 0,  0,  1}{blue} fonts indicate the first, second and third place, respectively. All results are generated solely by CPU.}
		\begin{tabular}{l c c c c c c c c c c c c }
			\hline
			& \textbf{ARCF-H}&\textbf{ARCF-HC}&{ECO-HC}&{STRCF}&{MCCT-H}&{STAPLE\_CA}&{SRDCF}&{BACF}&{MUSTER}&{SAMF}&{DSST}&{KCF}\\ \hline
			\textbf{FPS} & {51.2} & {15.3} & {41.1} &{22.6}&{32.1}&{37.2}&{11.7}&\textcolor[rgb]{ 0,  0,  1}{52.5}&{2.1}&{9.9}&\textcolor[rgb]{ 0,  1,  0}{100.7}&\textcolor[rgb]{ 1,  0,  0}{326.1}\\ 
			\textbf{MSPF} & {19.53} & {65.36} & {24.33} &{44.25}&{31.15}&{26.88}&{85.47}&\textcolor[rgb]{ 0,  0,  1}{19.05}&{476.19}&{101.01}&\textcolor[rgb]{ 0,  1,  0}{9.93}&\textcolor[rgb]{ 1,  0,  0}{3.07}\\ 
			\hline
		\end{tabular}%
		\label{tab:fps}%
	\end{table*}%
\end{center}

\subsection{Update of appearance model}
The appearance model $\mathbf{\hat{x}}^{\text{model}}$ is updated as follows:
\begin{equation}\label{update_strategy}
\mathbf{\hat{x}}^{\text{model}}_{k} = (1-\eta)\mathbf{\hat{x}}^{\text{model}}_{k-1} + \eta\mathbf{\hat{x}}_{k} \ ,
\end{equation}
\noindent where $k$ and $k-1$ denote $k$th and $(k-1)$th frame respectively. $\eta$ is the learning rate of the appearance model.

\section{Experiments}

In this section, the proposed ARCF tracker is exhaustively evaluated on 243 challenging image sequences with altogether over 90,000 frames from three widely applied benchmarks captured by UAV for tracking, respectively UAV123@10fps~\cite{Mueller2016ECCV}, DTB70~\cite{Li2017AAAI} and UAVDT~\cite{Du2018ECCV}. The results are compared with 20 state-of-the-art trackers with both hand-crafted based trackers and deep-based trackers, i.e., KCF~\cite{Henriques2015PAMI}, DSST~\cite{Danelljan2017PAMI}, SAMF~\cite{Li2014ECCV}, MUSTER~\cite{Hong2015CVPR}, BACF~\cite{Kiani2017ICCV}, SRDCF~\cite{Danelljan2015ICCV}, STAPLE\_CA~\cite{Mueller2017CVPR}, MCCT-H~\cite{Wang2018CVPR}, STRCF~\cite{Li2018CVPR}, ECO-HC (with gray-scale)~\cite{Danelljan2017CVPR}, ECO~\cite{Danelljan2017CVPR}, C-COT~\cite{Danelljan2016ECCV}, HCF~\cite{Ma2015ICCV}, ADNet~\cite{Yun2017CVPR}, CFNet~\cite{Valmedre2017CVPR}, CREST~\cite{Song2017ICCV}, MCPF~\cite{Zhang2017CVPR},  SINT~\cite{Tao2016CVPR}, SiamFC~\cite{Bertinetto2016ECCV}, and HDT~\cite{Qi2016CVPR}. All evaluation criteria are according to the original protocol defined in three benchmarks respectively~\cite{Mueller2016ECCV, Li2017AAAI, Du2018ECCV}.

%

\subsection{Implementation details}
\begin{figure}[!t]
	\begin{center}
		\centering
		\includegraphics[width=1\columnwidth]{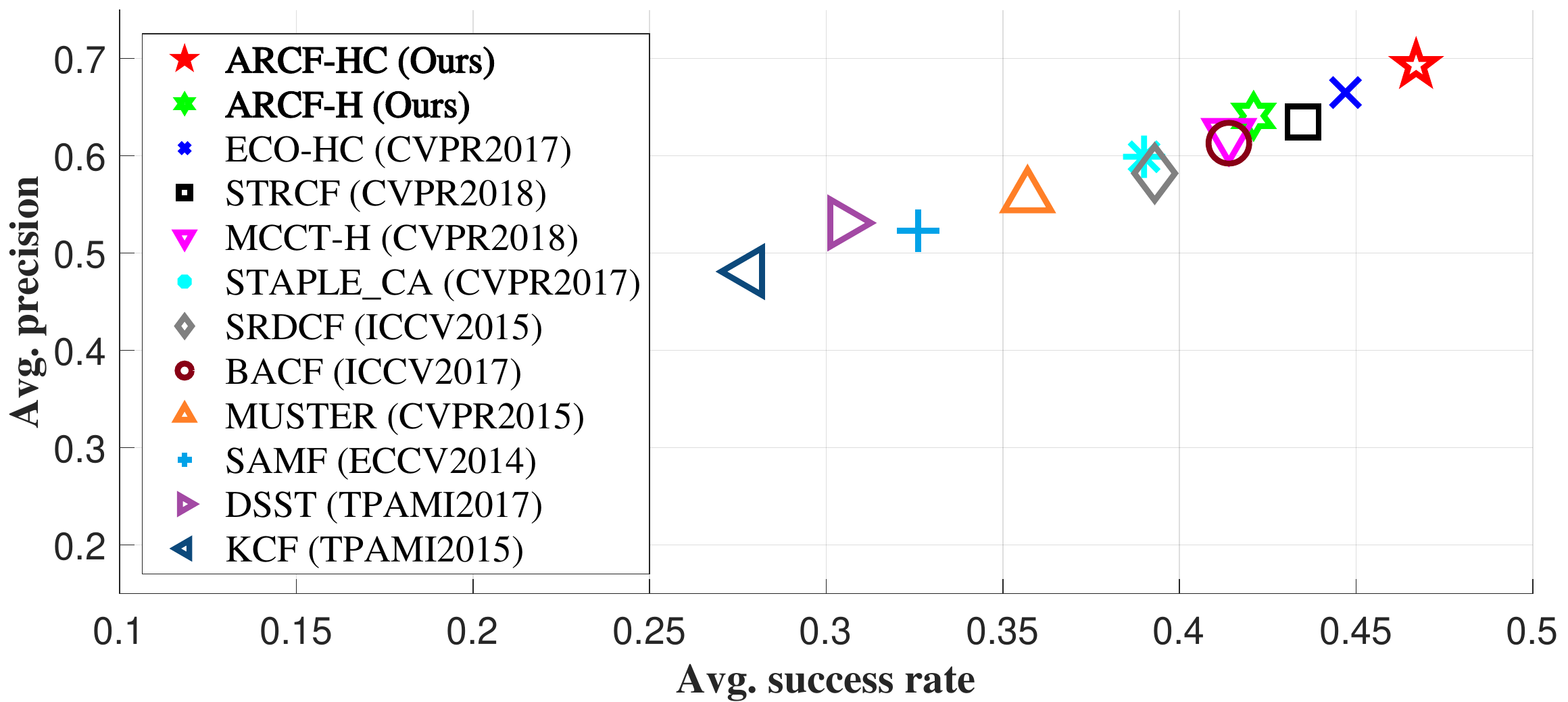}
		\caption{Comparison of different state-of-the-art trackers based on hand-crafted features. Value of average precision and average success rate is calculated by averaging OPE results from three datasets. }
		\label{fig:compscat}
	\end{center}
\end{figure}
\begin{figure*}[!t]
	\begin{center}

		\subfigure[] { \label{fig:at_a} 
			\begin{minipage}{0.32\textwidth}
				\centering
				\includegraphics[width=1\columnwidth]{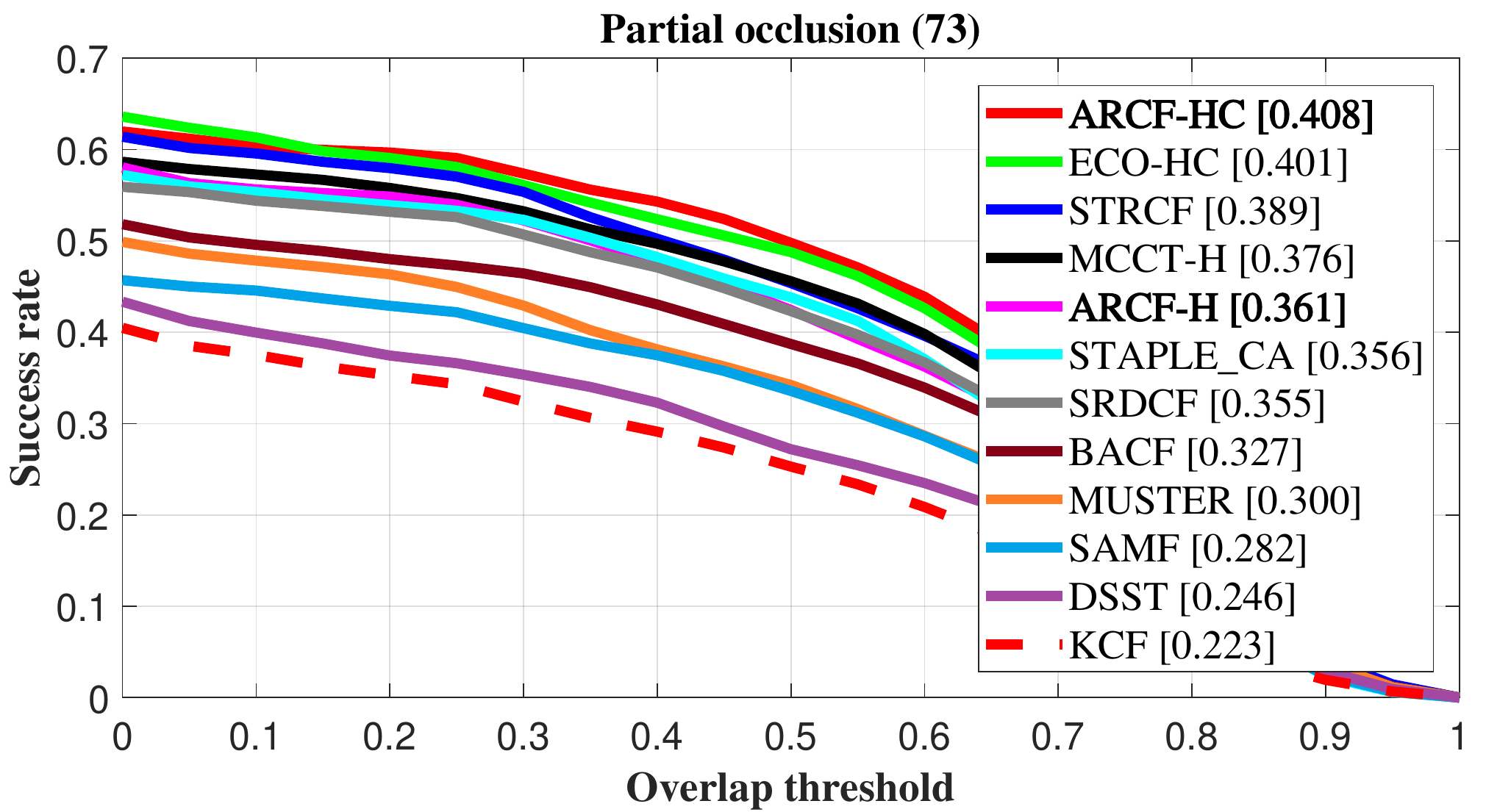}
				\\
				\includegraphics[width=1\columnwidth]{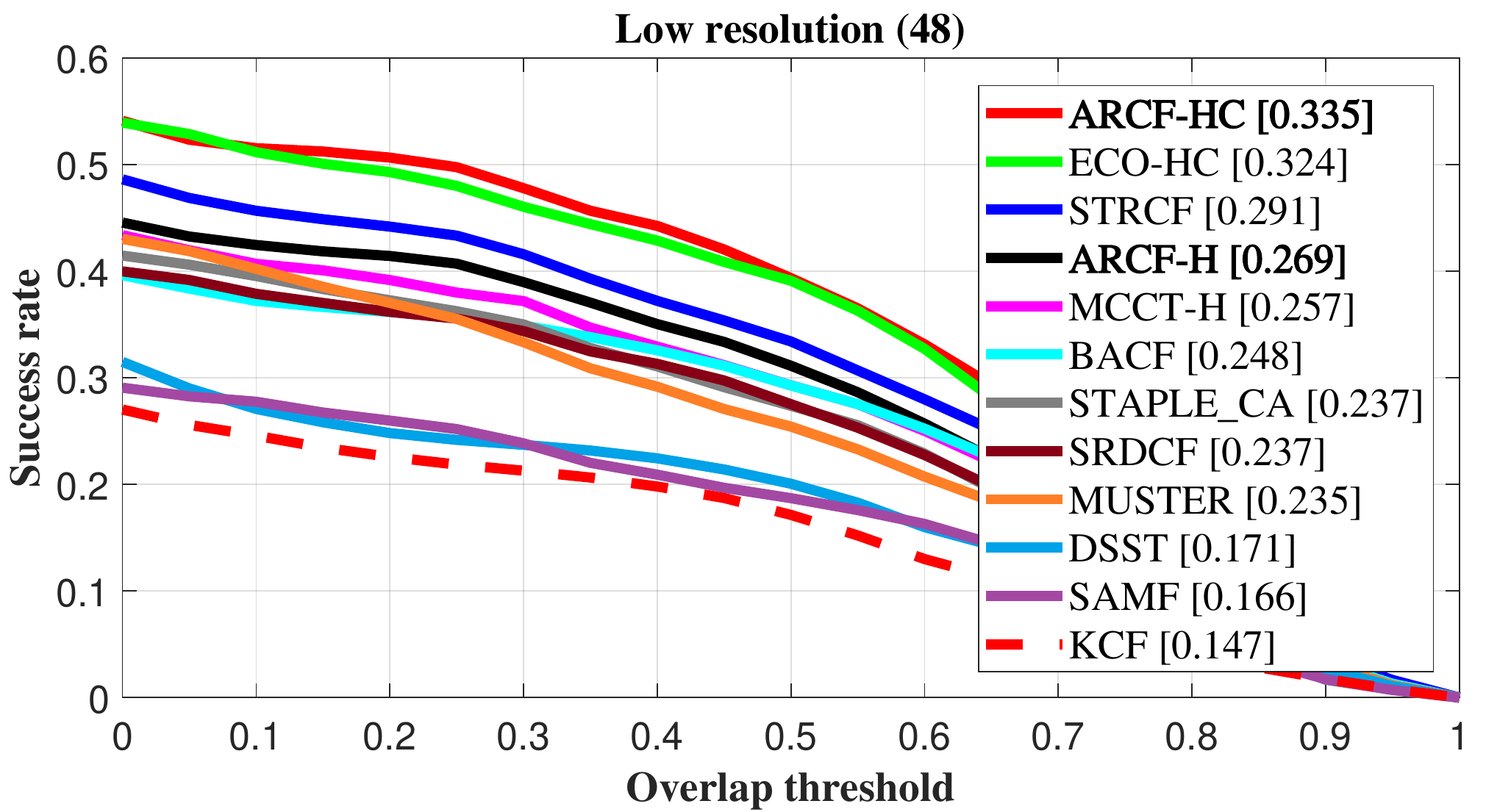}
			\end{minipage}
		}
		\subfigure[] { \label{fig:at_b} 
			\begin{minipage}{0.32\textwidth}
				\centering
				\includegraphics[width=1\columnwidth]{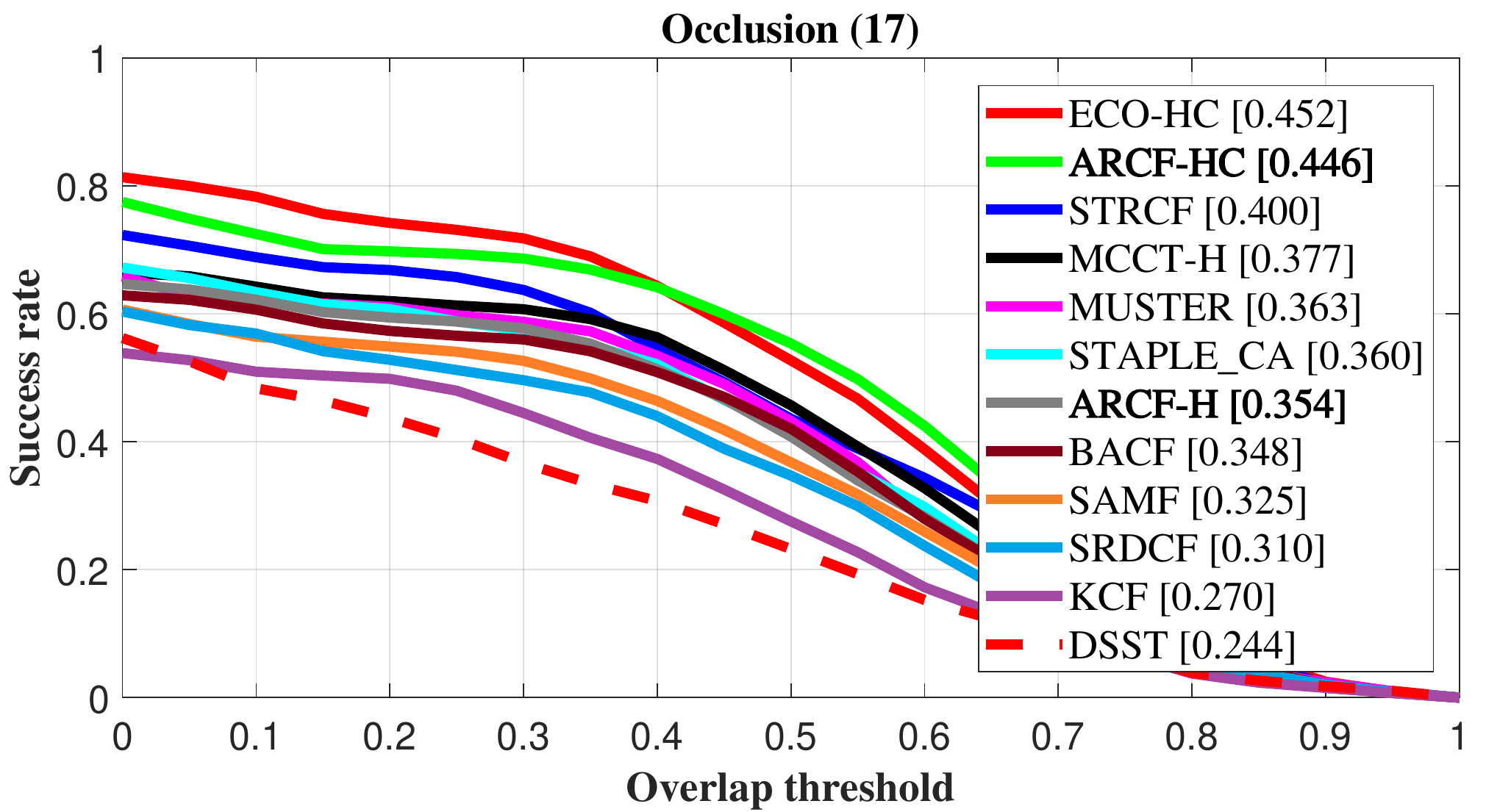}
				\\
				\includegraphics[width=1\columnwidth]{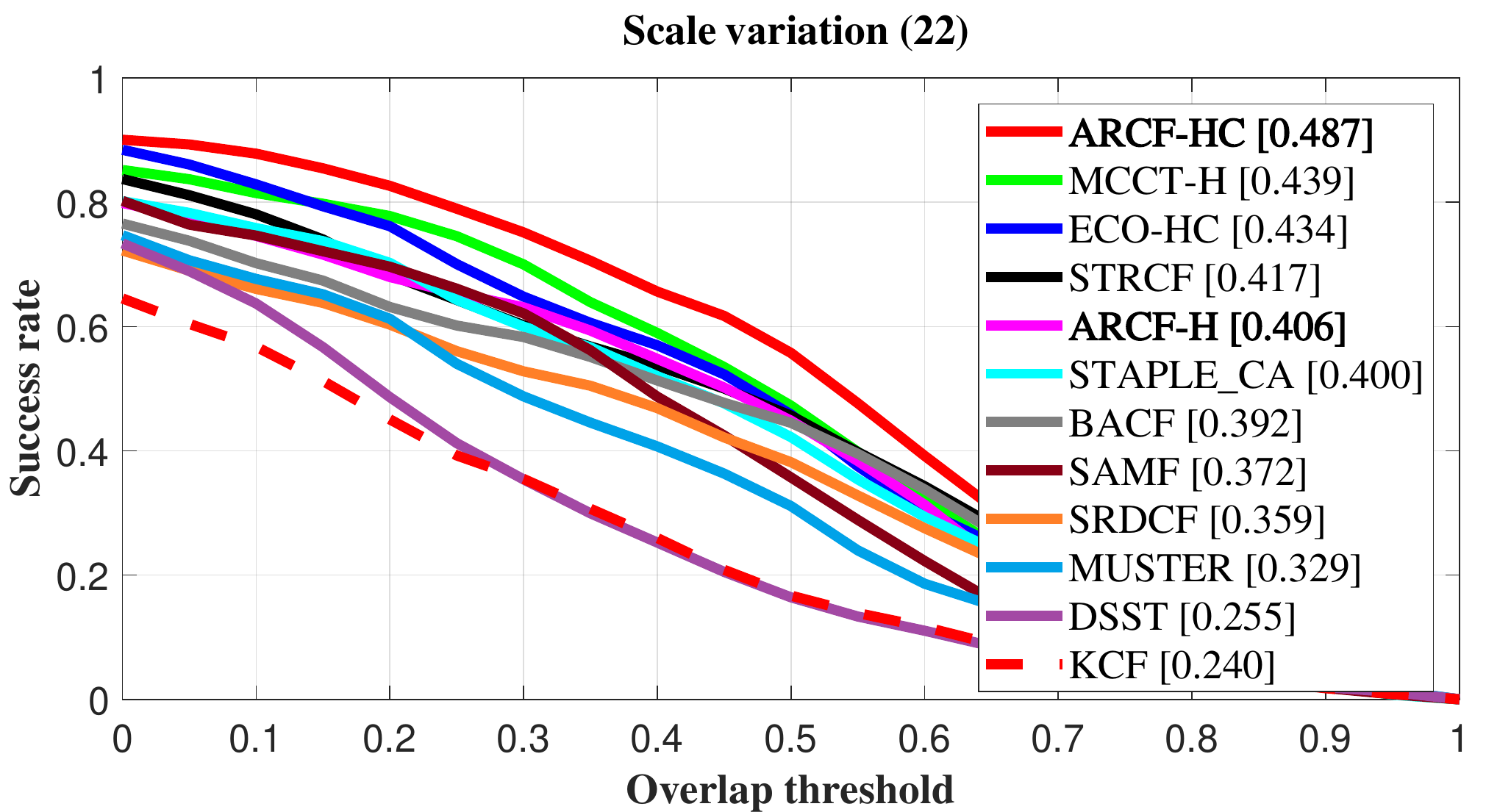}
			\end{minipage}
		}
		\subfigure[] { \label{fig:at_c} 
			\begin{minipage}{0.32\textwidth}
				\centering
				\includegraphics[width=1\columnwidth]{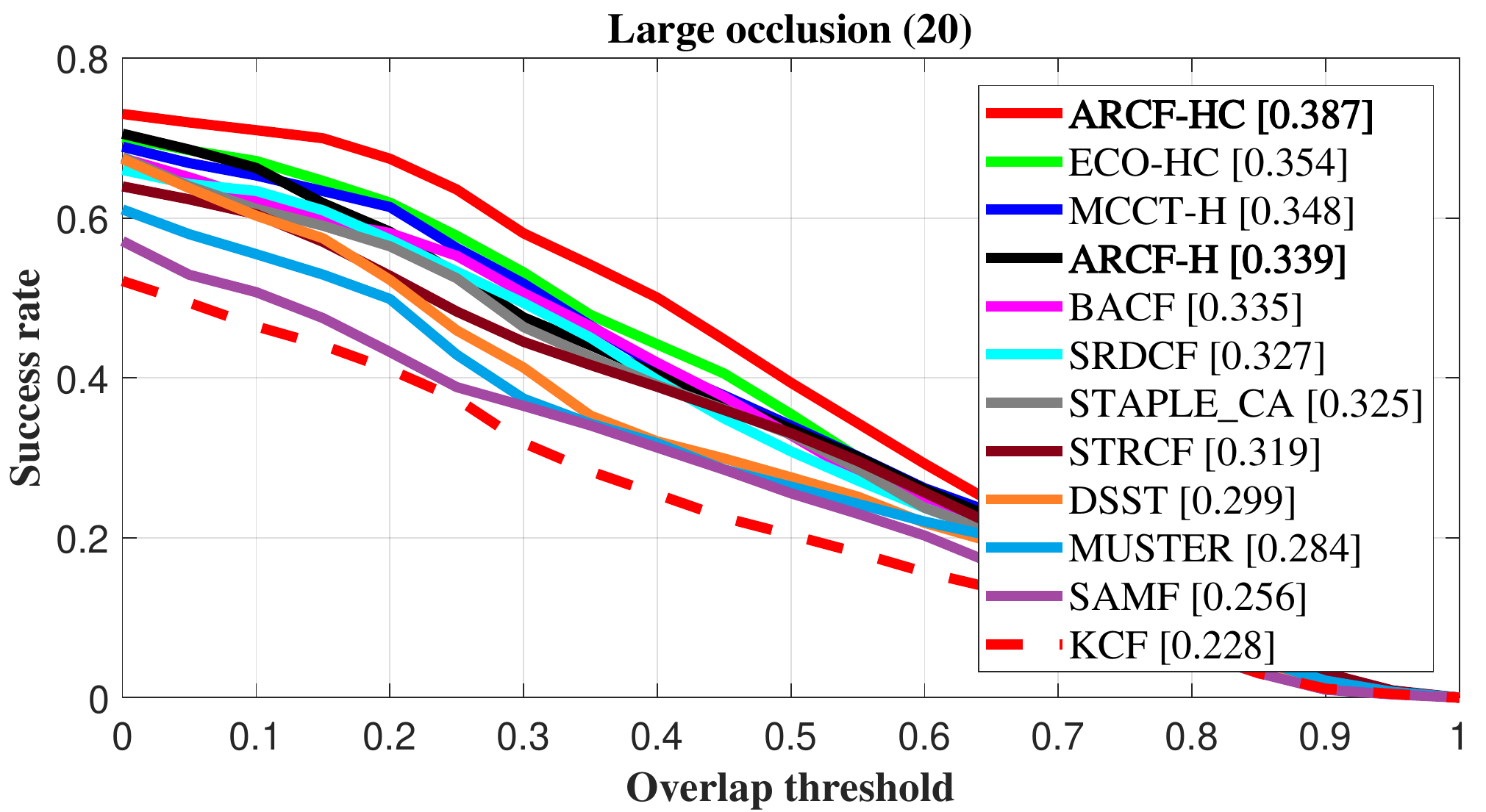}
				\\
				\includegraphics[width=1\columnwidth]{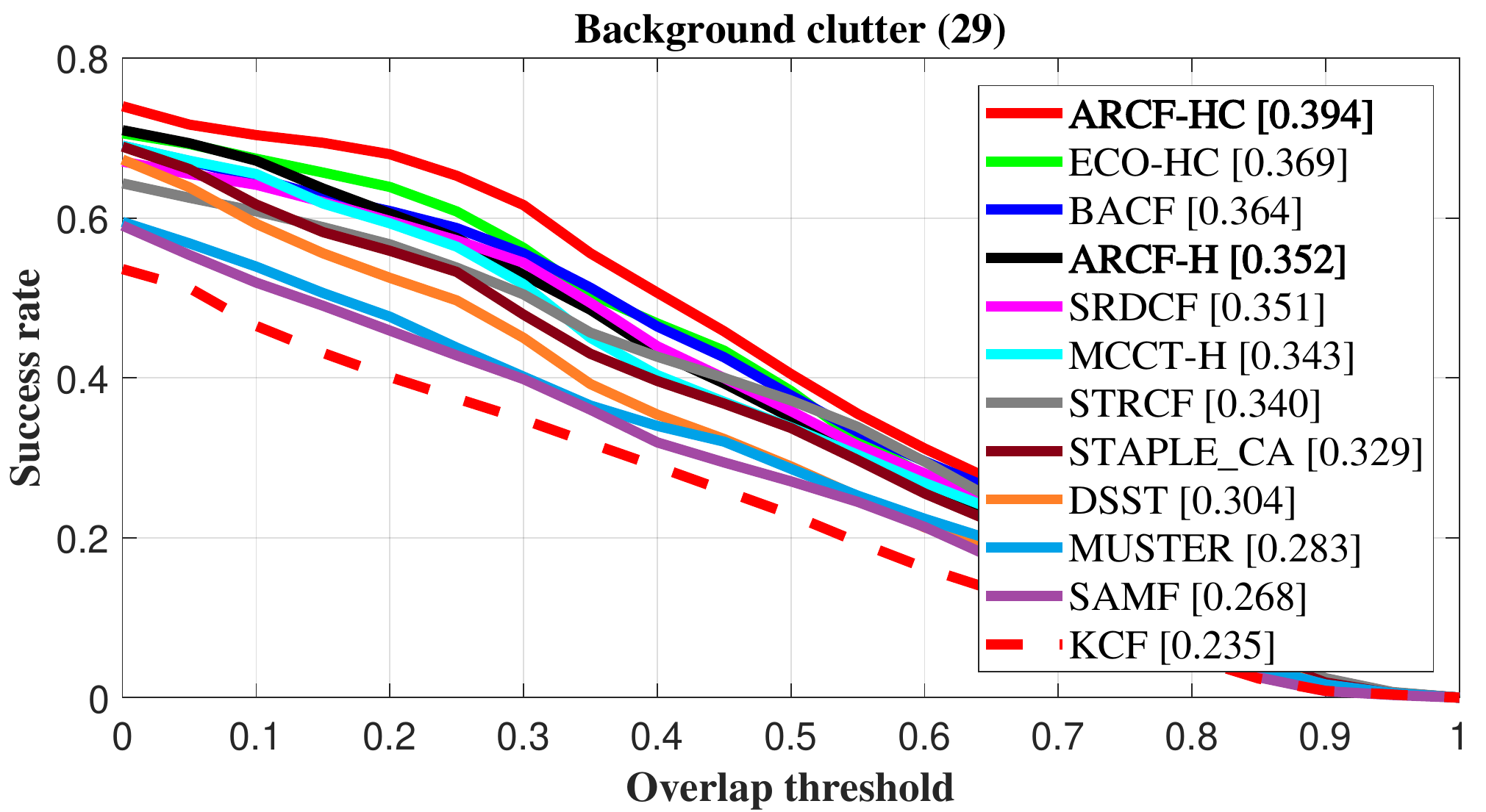}
			\end{minipage}
		}
	\end{center}
	
	\caption{Attribute based evaluation. Success plots of attributes comparing ARCF-HC and ARCF-H with other state-of-the-art hand-crafted based trackers on (a) UAV123@10fps, (b) DTB70 and (c) UAVDT. AUC is used to rank different trackers. Detailed definitions and descriptions of these attributes can be seen in~\cite{Mueller2016ECCV, Li2017AAAI, Du2018ECCV}. }
	\label{fig:at}
\end{figure*}
\urlstyle{rm}
Two versions of ARCF tracker, respectively ARCF-H (with only HOG feature) and ARCF-HC (with HOG, CN and gray-scale features) are developed in the experiment to achieve comprehensive comparison with all trackers using HOG, both HOG and CN, as well as deep features. 
The value of $\gamma$ is set to $0.71$, ADMM iteration is set to $5$ and the learning rate  $\eta$ is $0.0192$. All experiments of all 21 trackers are carried out by MATLAB R2017a on a computer with an i7-8700K processor (3.7GHz), 48GB RAM and NVIDIA Quadro P2000 GPU. Tracking code is available here: \url{https://github.com/vision4robotics/ARCF-tracker}.

\subsection{Comparison with hand-crafted based trackers}

\subsubsection{Quantitive evaluation}
\textbf{Overall performance evaluation:} Figure \ref{fig:hc} demonstrates the overall performance of ARCF-H and ARCF-HC with other state-of-the-art hand-crafted feature-based trackers on UAV123@10fps, DTB70 and UAVDT datasets. The proposed ARCF-HC tracker has outperformed all other trackers based on hand-crafted features on all three datasets. More specifically, on UAV123@10fps dataset, ARCF-HC (0.666) has an advantage of 0.6\% and 3.9\% over the second and third best tracker ECO-HC (0.660), STRCF (0.627) respectively in precision, as well as an advantage of 0.1\% and 1.6\% over the second (ECO-HC, 0.472) and third best tracker (STRCF, 0.457) respectively in AUC. On DTB70 dataset, ARCF-HC (0.694, 0.472) also achieved the best performance, followed by ECO-HC (0.648, 0.457) and STRCF (0.649, 0.437). On UAVDT, ARCF-HC tracker (0.720, 0.458) is closely followed by ARCF-H (0.705) and BACF (0.432) in precision and AUC respectively. Overall evaluation of performance on all three datasets in terms of precision and AUC is demonstrated in Fig.~\ref{fig:compscat}. Against the baseline BACF, ARCF-H has an advancement of 2.77\% in precision  and 0.69\% in AUC. ARCF-HC has made a progress of 7.98\% and 5.32\% in precision and AUC respectively. Besides satisfactory tracking results, the speed of ARCF-H and ARCF-HC is adequate for real-time UAV tracking applications, as shown in Table~\ref{tab:fps}.

\textbf{Attribute based comparison:} In this section, quantitative analysis of different attributes in three benchmarks are performed. The proposed ARCF-HC tracker has performed favorably against other top hand-crafted based trackers in most attributes defined respectively in three benchmarks. Examples of overlap success plots are demonstrated in Fig.~\ref{fig:at}. In partial or full occlusion cases, ARCF-H and ARCF-HC demonstrated a huge improvement from its baseline BACF, and have achieved state-of-the-art performance in this aspect on all three benchmarks. Usually, in occlusion cases, CF learns appearance model of both the tracked object and irrelevant objects that caused occlusions. ARCF is able to restrict the learning of irrelevant objects by restricting the response map variations, thus achieving a better performance in occlusion cases. More specifically, ARCF-HC has achieved an advancement of 8.1\% (UAV123@10fps), 9.8\% (DTB70) and 5.2\% (UAVDT) respectively from BACF in AUC in occlusion cases. In other attributes, ARCF-H and ARCF-HC have also shown a great improvement from BACF and achieved a performance with a high ranking. More complete results of attribute evaluation can be found on supplementary materials.
\begin{figure}[!t]
	\begin{center}
		\includegraphics[width=1\columnwidth]{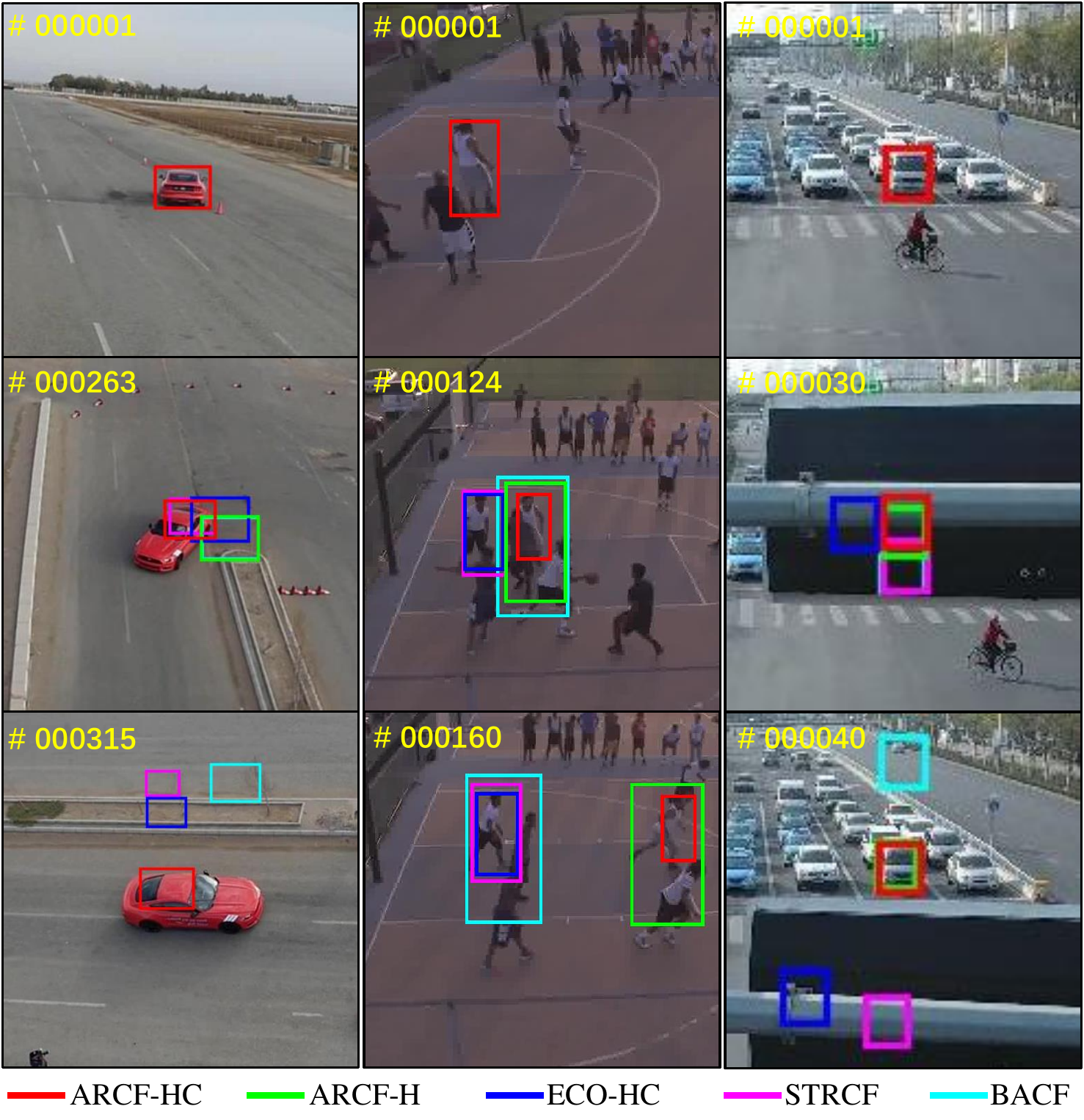}
		\caption{Qualitative performance evaluation of ARCF-H and ARCF-HC tracker on \textit{Car16\_2} from UAV123 dataset, \textit{StreetBasketball3} from DTB70 dataset and \textit{S0601} from UAVDT dataset.}
		\label{fig:qualitative_results}
	\end{center}
\end{figure}
\subsubsection{Qualitative evaluation}
Some qualitative tracking results of ARCF and other top trackers are shown in Fig.~\ref{fig:qualitative_results}. It can be proven that ARCF is competent in dealing with both partial as well as full occlusions and performs satisfactorily in other aspects defined in three benchmarks as well.
\begin{figure}[!t]
	\begin{center}
		\begin{minipage}{0.38\textwidth}
			\centering
			\includegraphics[width=1\columnwidth]{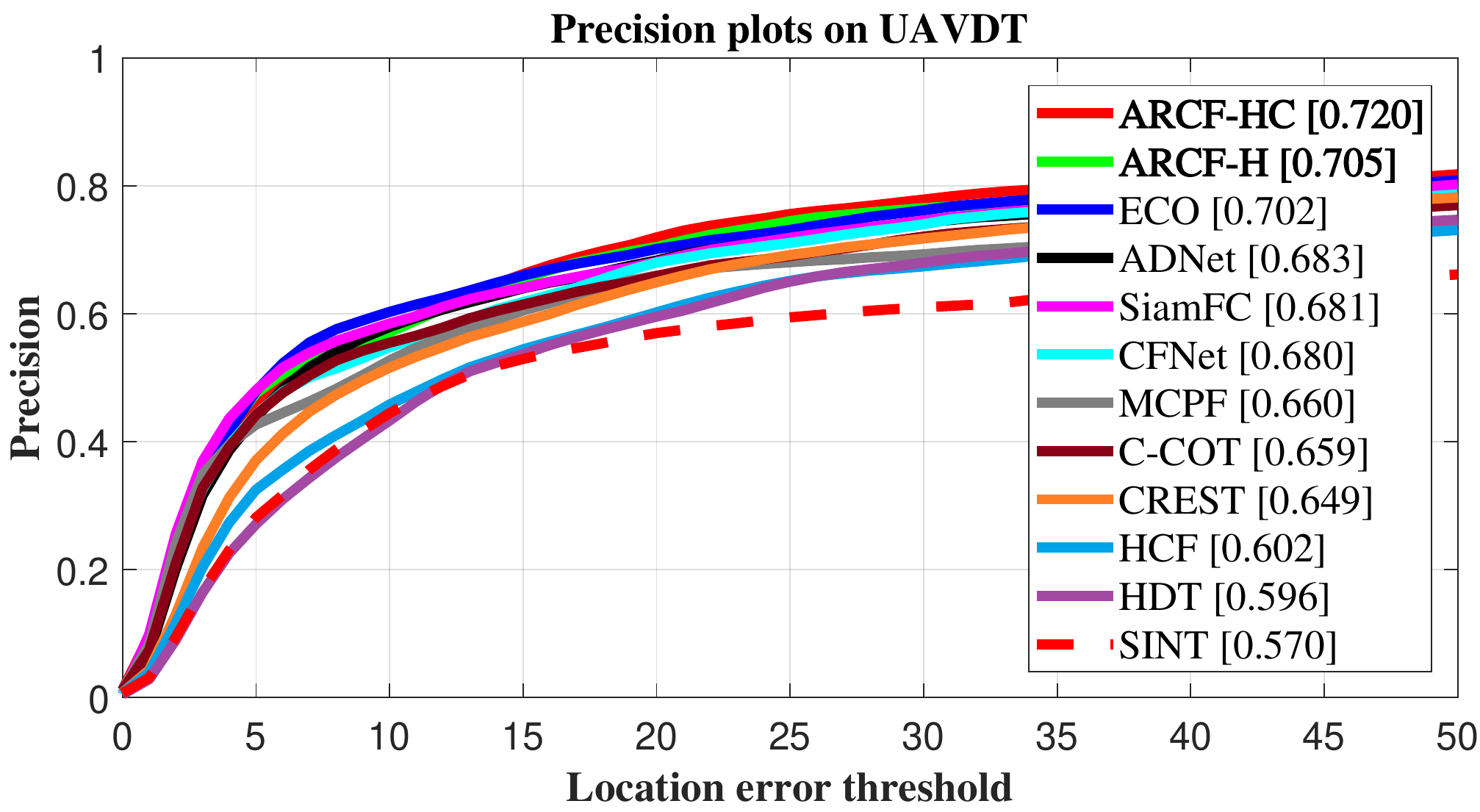}
			\\
			\includegraphics[width=1\columnwidth]{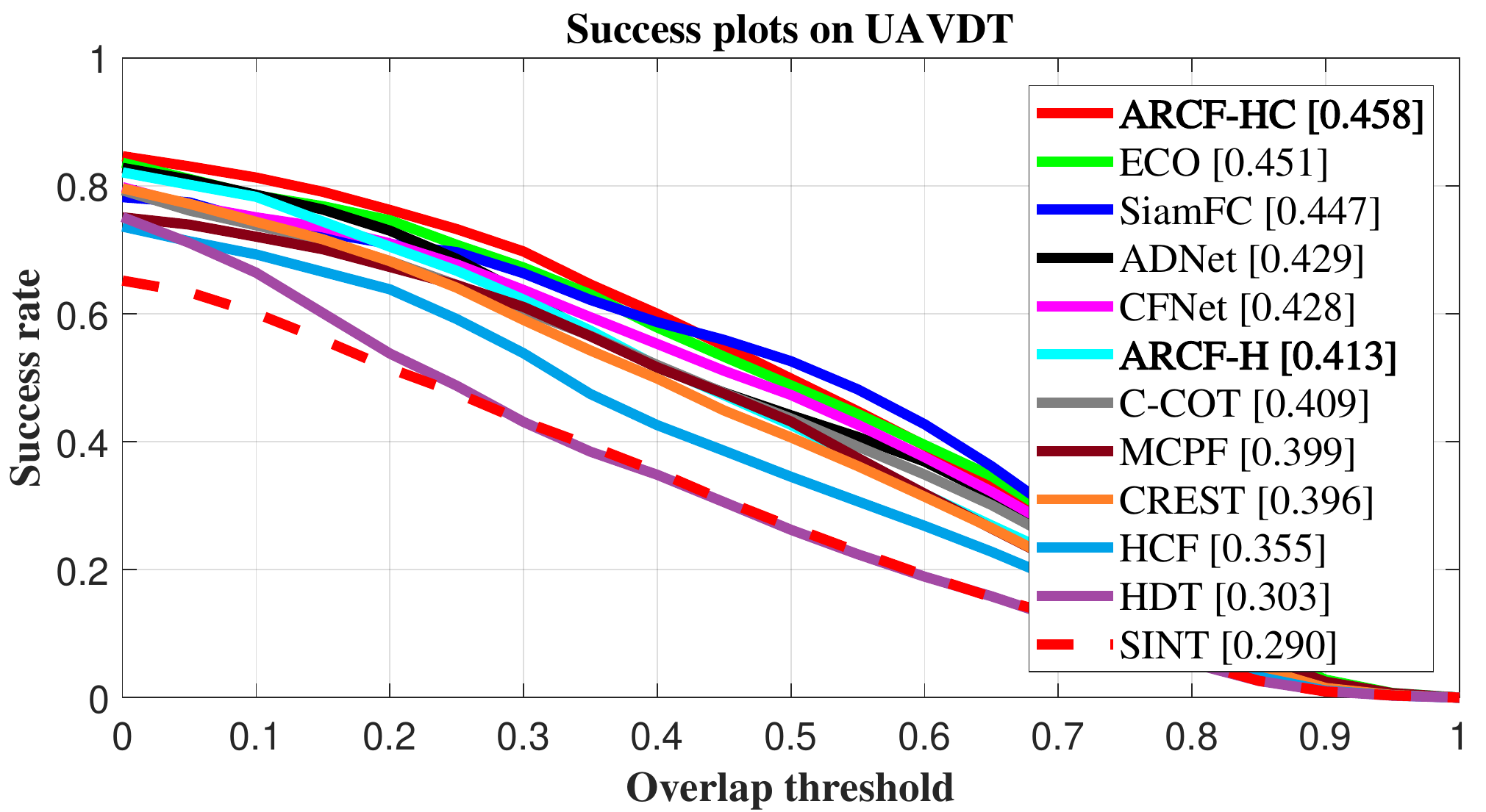}
		\end{minipage}
		\caption{Comparison between ARCF tracker and different state-of-the-art deep-based trackers. Value of average precision and average success rate is calculated by averaging OPE results from three datasets. }
		\label{fig:deep_comparision}
	\end{center}
\end{figure}
\subsection{Comparison with deep-based trackers}
To achieve a more comprehensive evaluation of the proposed trackers ARCF-H and ARCF-HC, these two trackers are also compared to ones using deep features or even deep trackers. In terms of precision and success rate, ARCF-HC has also performed favorably against other state-of-the-art deep-based trackers. Fig.~\ref{fig:deep_comparision} has shown the quantitative comparison on UAVDT dataset. 
\begin{center}
	\begin{table}[!t]
		\small
		\setlength{\tabcolsep}{3mm}
		\centering
		\caption{Average map difference comparison of BACF and ARCF-H on different datasets. Map difference is evaluated by Eq.~\ref{eq:difference}. \textbf{Bold} font indicates lower average difference.}
		\begin{tabular}{l c c c}
			\hline
			& UAV123@10fps&DTB70&UAVDT\\ \hline\hline
			ARCF-H & \textbf{0.0106} & \textbf{0.0098} & \textbf{0.0074}
			\\
			BACF & {0.0133} & {0.0129} & {0.0087}\\ \hline
		\end{tabular}%
		\label{tab:avg_map_diff}%
	\end{table}%
\end{center}
\begin{figure}[!t]
	\begin{center}
		\includegraphics[width=1\linewidth]{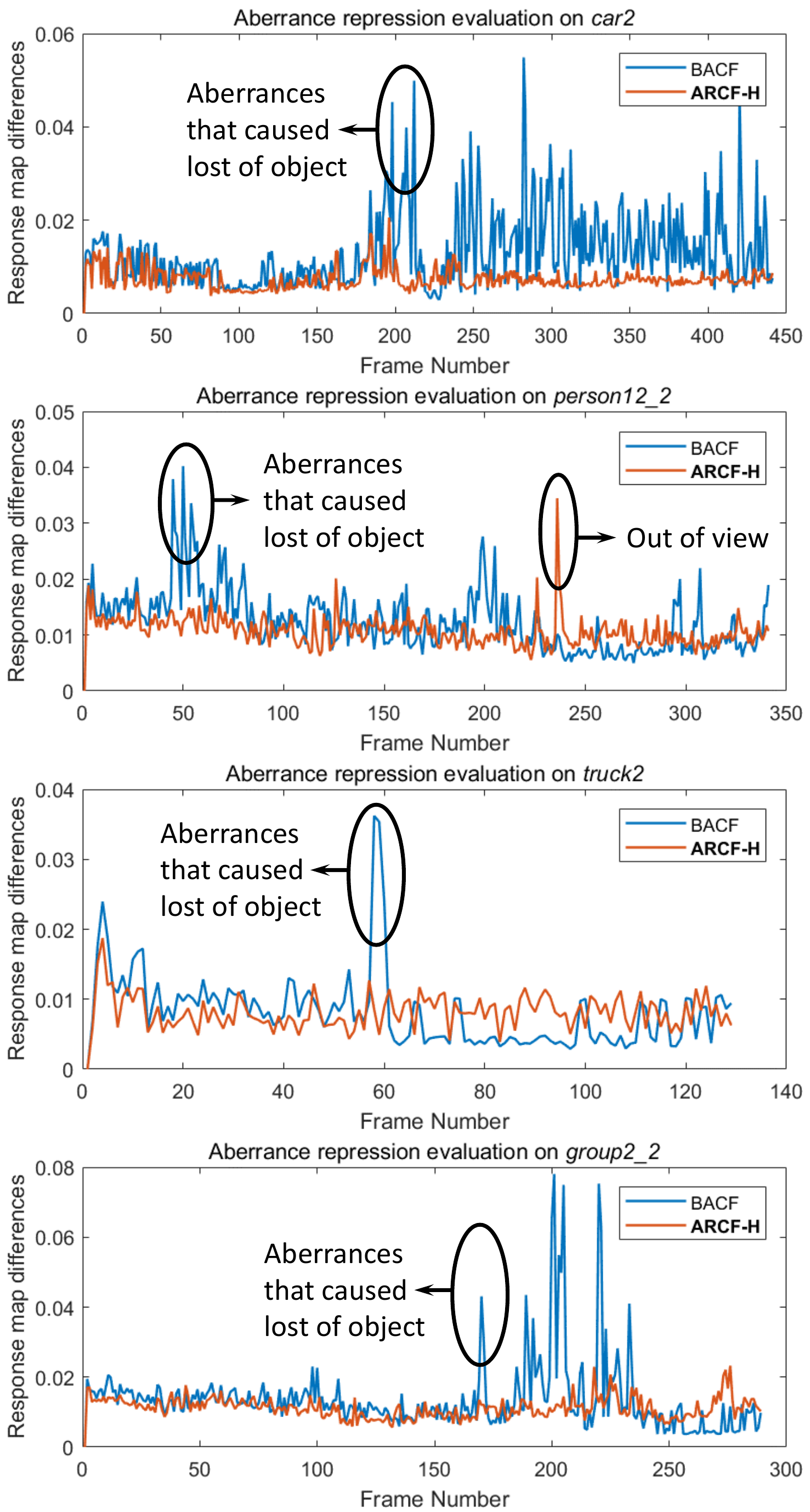}
	\end{center}
	\caption{Comparison of response map differences between BACF tracker and ARCF tracker on UAV123@10fps dataset, specifically on \textit{car2}, \textit{person12\_2}, \textit{truck\_2} and \textit{group2\_2}. The proposed ARCF tracker has remarkably repressed aberrances that can possibly cause lost of object. Note that after the out-of-view on \textit{person12\_2}, ARCF rapidly recapture the original tracked object. }
	\label{fig:aberrance_repression_evaluation}
\end{figure}
\subsection{Aberrance repression evaluation}
\label{sec:aberrance_repression_evaluation}
In order to illustrate the effect of aberrance repression, this section investigates the difference between tracking performance of BACF and ARCF-H trackers. It can be clearly seen from Table \ref{tab:avg_map_diff} that ARCF-H tracker has significantly repressed the average map difference compared to BACF for respectively 20\%, 24\%, and 15\% on UAV123@10fps, DTB70 and UAVDT dataset. Response map differences are visualized in Fig. \ref{fig:aberrance_repression_evaluation} to demonstrate the performance of aberrrance repression method. When objects go through relatively big appearance changes due to sudden illumination variation, partial or full occlusion and other reasons, response map tends to fluctuate and aberrances are very likely to happen, as denoted in Fig.~\ref{fig:aberrance_repression_evaluation}. Although it is possible in cases like out-of-view and full occlusion that aberrances happen in ARCF tracker, ARCF is able to suppress most undesired fluctuations so that the tracker can be more robust against these appearance changes. It should be brought to attention that this kind of fluctuation is omnipresent in tracking scenarios of various image sequences. More examples of visualization of response map differences can be seen in the supplementary material.

%

\section{Conclusion and future work}
In this work, aberrance repressed correlation filters have been proposed for UAV visual tracking. By introducing a regularization term to restrict the response map variations to BACF, ARCF is capable of suppressing aberrances that is caused by both background noise information introduced by BACF and appearance changes of the tracked objects. After careful and exhaustive evaluation on three prevalent tracking benchmarks captured by UAVs, ARCF has proved itself to have achieved a big advancement from BACF and have state-of-the-art performance in terms of precision and success rate. Its speed is also more than sufficient for real-time UAV tracking. In conclusion, the proposed method i.e., aberrance repression correlation filters (ARCF), is able to raise the performance of DCF trackers without sacrificing much speed. Out of consideration for computing efficiency due to application of UAV tracking, the proposed ARCF has only used HOG and CN as extracted feature. In cases with low demand for real-time application, more comprehensive features such as convolutional ones can be applied to ARCF for better precision and success rate. Also, the framework of aberrance repression can be extended to other trackers like ECO~\cite{Danelljan2017CVPR} and SRDCF~\cite{Danelljan2015ICCV}. We believe, with our proposed aberrance repression method, DCF framework and the performances of DCF based trackers can be further improved.

\noindent\textbf{Acknowledgment: }This work is supported by the National Natural Science Foundation of China (No. 61806148) and the Fundamental Research Funds for the Central Universities (No. 22120180009).

{\small
\bibliographystyle{ieee_fullname}
\bibliography{reference}
}

\end{document}